\documentclass[10pt,twocolumn,letterpaper]{article}

\usepackage{cvpr}              % To produce the CAMERA-READY version

\usepackage{booktabs} % For professional-looking tables
\usepackage{multirow} % To span cells over multiple rows
\usepackage{graphicx} % To resize the table if needed
\usepackage{amsmath}  % For math symbols like \delta
\usepackage[dvipsnames]{xcolor}
\usepackage{comment}
\usepackage{caption}

\newcommand{\catA}{\raisebox{-0.4ex}{\textcolor{RoyalBlue}{{\Large\textbullet}}}}
\newcommand{\catB}{\raisebox{-0.4ex}{\textcolor{BurntOrange}{{\Large\textbullet}}}}
\newcommand{\catC}{\raisebox{-0.4ex}{\textcolor{OliveGreen}{{\Large\textbullet}}}}
\newcommand{\catD}{\raisebox{-0.4ex}{\textcolor{RedViolet}{{\Large\textbullet}}}}
\newcommand{\catE}{\raisebox{-0.4ex}{\textcolor{TealBlue}{{\Large\textbullet}}}}
\newcommand{\catF}{\raisebox{-0.4ex}{\textcolor{Brown}{{\Large\textbullet}}}}

\newcommand{\xy}[1]{{\color{blue}{xy:}#1}}
\newcommand{\syt}[1]{\textcolor[rgb]{1,0.4,0.0}{#1}}
\newcommand{\xjqi}[1]{\textcolor[rgb]{1,0,0}{{[{xjqi:} #1]}}}

\definecolor{cvprblue}{rgb}{0.21,0.49,0.74}
\usepackage[pagebackref,breaklinks,colorlinks,allcolors=cvprblue]{hyperref}
\usepackage{algorithmicx}
\usepackage{algpseudocode}
\usepackage{cuted}
\usepackage{adjustbox}
\usepackage{multirow}
\usepackage{tabularx}
\usepackage{pifont}
\usepackage{makecell}

\def\paperID{} % *** Enter the Paper ID here
\def\confName{CVPR}
\def\confYear{2026}

\title{Stabilizing Streaming Video Geometry via Dynamic Feature Normalization}

\vspace{-10pt}
\author{
\begin{tabular}{cccc}
Xiaoyang Lyu$^{*1}$ & Muxin Liu$^{*1}$ & Xiaoshan Wu$^1$ & Ruicheng Wang$^2$ \\
Yi-Hua Huang$^1$ & Yang-Tian Sun$^1$ & Shaoshuai Shi$^3$ & Xiaojuan Qi$^{1 \diamond}$
\end{tabular} \\[2ex]
{\normalsize $^1$The University of Hong Kong \quad $^2$USTC \quad $^3$Voyager Research, Didi Chuxing} \\
{\tt\footnotesize $^*$ Equal Contribution: \{shawlyu, mxliu\}@connect.hku.hk}  
{\tt\footnotesize $^\diamond$ Corresponding Author: xjqi@eee.hku.hk}
}

\begin{document}
\maketitle

\begin{strip}
     \centering
     \vspace{-18mm}
     \includegraphics[width=\linewidth]{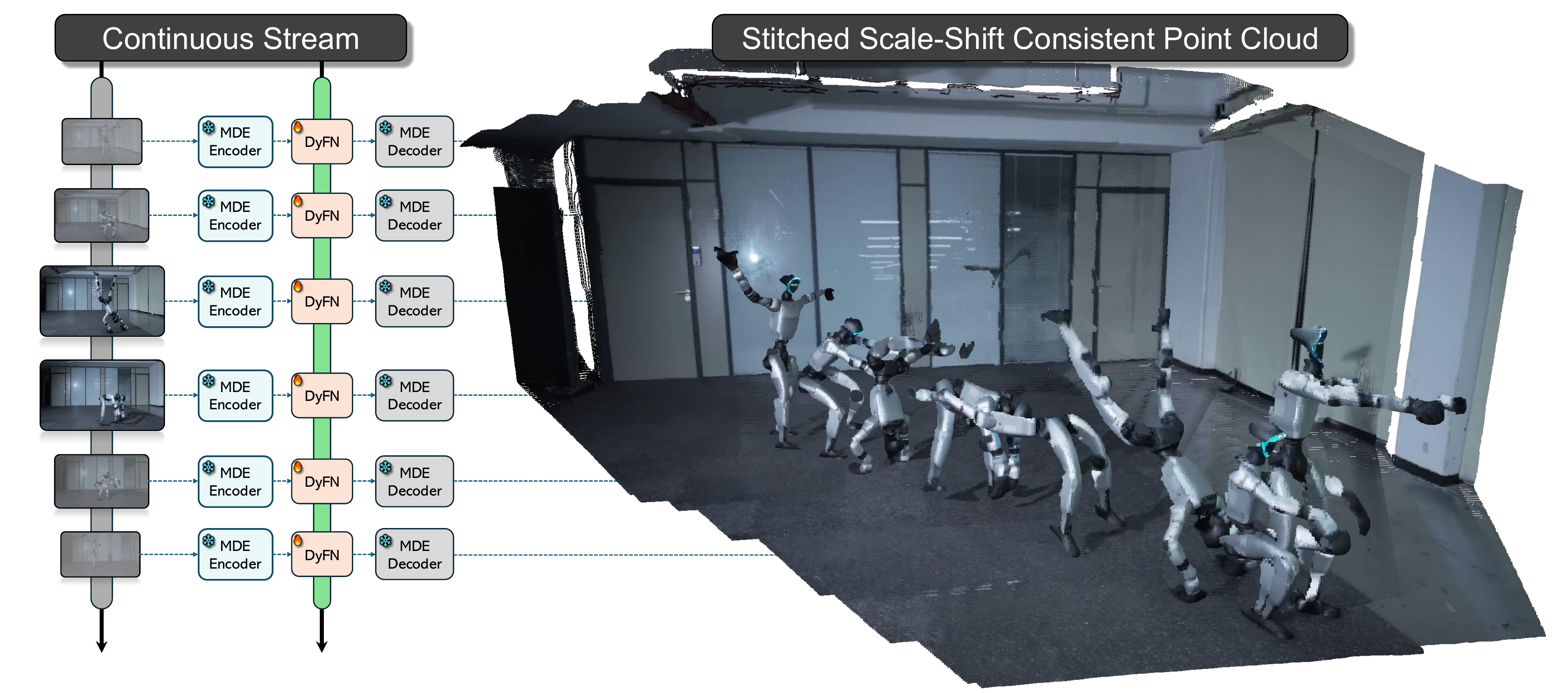}
     \vspace{-6mm}
     \captionsetup{hypcap=false}\captionof{figure}{We introduce Dynamic Feature Normalization (DyFN), a novel module that transforms single-image geometry estimators into consistent streaming video models. DyFN achieves scale-shift consistency across continuous frames, enabling the generation of coherent 3D reconstructions from streaming monocular input.}
     \label{fig:teaser}
     \vspace{-2mm}
\end{strip}

\begin{abstract}
Consistent 3D geometry estimation from streaming RGB input is crucial for real-world applications such as autonomous driving, embodied AI, and large-scale reconstruction. 
While modern monocular geometry foundation models achieve strong single-image accuracy, they exhibit severe temporal inconsistency on continuous input, notably dominated by scale–shift drifting. Through targeted empirical analysis, we trace this instability to its root cause: fluctuations in latent feature statistics, whose mean and variance directly determine the predicted depth’s scale and shift. Building on this insight, we introduce Dynamic Feature Normalization (DyFN), a lightweight, causal recurrent module that dynamically and robustly modulates feature statistics to maintain stable geometry over time. We adapt powerful pretrained monocular geometry models for streaming by finetuning only DyFN, a mere 2\% additional parameters, while keeping the backbone frozen, thereby achieving temporal consistency without compromising single-image accuracy. Extensive experiments across four benchmarks show that DyFN effectively eliminates temporal artifacts such as disjointed layering and positional jitter, and achieves state-of-the-art temporal stability, improving over prior streaming methods by up to 14\% and even outperforming heavier non-causal video baselines. Project page: \small\url{https://shawlyu.github.io/DyFN}
\end{abstract}
\section{Introduction}
\label{sec:intro}
3D geometry estimation is fundamental to many real-world applications, such as robotics, autonomous driving, and augmented reality. 
Recently, Monocular Geometry Estimation (MGE) and Monocular Depth Estimation (MDE)~\cite{wang2025moge,  wang2025moge2,yang2024depth, yang2024depthv2, bhat2023zoedepth, ranftl2020towards, yin2023metric3d, piccinelli2024unidepth} has progressed rapidly with the rise of large-scale foundation models, significantly narrowing the gap between single-image prediction and sensor-based measurements. 
% Models such as MoGe~\cite{wang2025moge} have demonstrated impressive zero-shot generalization across diverse scenes by learning geometry-aware representations from massive image collections. 
Models such as MoGe~\cite{wang2025moge} further exhibit remarkable zero-shot generalization across diverse scenes by learning geometry-aware priors from massive image collections.
Despite these advances, most MGE and MDE models are designed for image inference, limiting their applicability in dynamic environments where input naturally arrives as continuous video streams. 

\begin{figure}[tbp]
  \centering
  \includegraphics[width=0.85\linewidth]{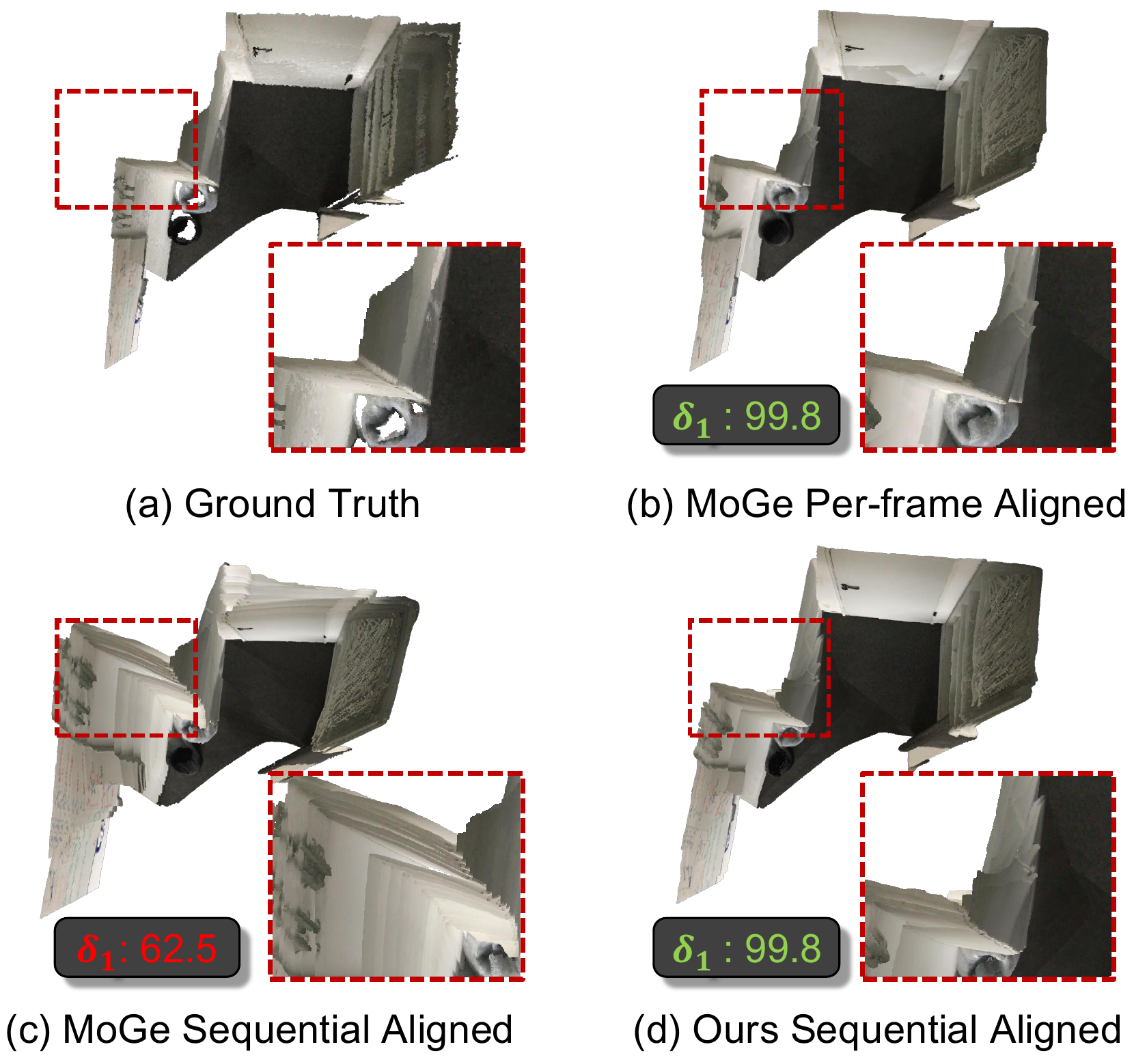}
    \vspace{-5pt}
  \caption{\textbf{Reconstruction comparison.} We align the predicted depth to metric scale using an affine transformation. Per-frame Aligned involves calculating the scale and shift for each frame independently. Sequence-aligned involves calculating a single, consistent scale and shift for the entire sequence. The point clouds are then fused using ground truth poses.($\delta_1$ means $\delta < 1.25$)}
  \label{fig:recon_evidence}
  \vspace{-12pt}
\end{figure}

% static and offline usage,
%while ours is dynamic and online/streaming
% When extended to continuous frame inputs, image-based MGE models exhibit severe temporal inconsistency, geometry predictions fluctuate across consecutive frames, leading to geometric distortions like disjointed layering and positional jitter in reconstructed scenes (Fig.~\ref{fig:recon_evidence} (c)). 
When applied to continuous video streams, image-based MGE models exhibit pronounced temporal inconsistency: geometry predictions fluctuate across frames, causing distortions such as layering breaks and positional jitter in reconstructed scenes (Fig.~\ref{fig:recon_evidence}c).
% Existing approaches attempt to alleviate this issue by introducing temporal attention mechanisms~\cite{hu2024depthcrafter, shao2024learningtemporallyconsistentvideo} or recurrent memory modules~\cite{chou2025flashdepth, wang2025continuous} to maintain inter-frame consistency.  
Existing methods mitigate this issue using temporal attention~\cite{hu2024depthcrafter, shao2024learningtemporallyconsistentvideo} or recurrent memory modules~\cite{chou2025flashdepth, wang2025continuous} to enforce inter-frame coherence.
% However, these approaches come with notable limitations. 
% First, they usually require full-network finetuning on large-scale, annotated video datasets, which is computationally expensive and data-hungry.  
% Second, full video finetuning often compromises the per-frame accuracy and zero-shot generalization ability of pretrained MGE models, as the backbone is overfitted to specific video domains and loses its original spatial precision. 
However, these solutions come with notable limitations:
they typically require full-network finetuning on large-scale annotated video datasets, which is computationally expensive and data-intensive; and
such finetuning often degrades the per-frame accuracy and zero-shot generalization of pretrained MGE models by overfitting the backbone to specific video domains.

\begin{figure}[tbp]
  \vspace{-5pt}
  \centering
  \includegraphics[width=0.95\linewidth]{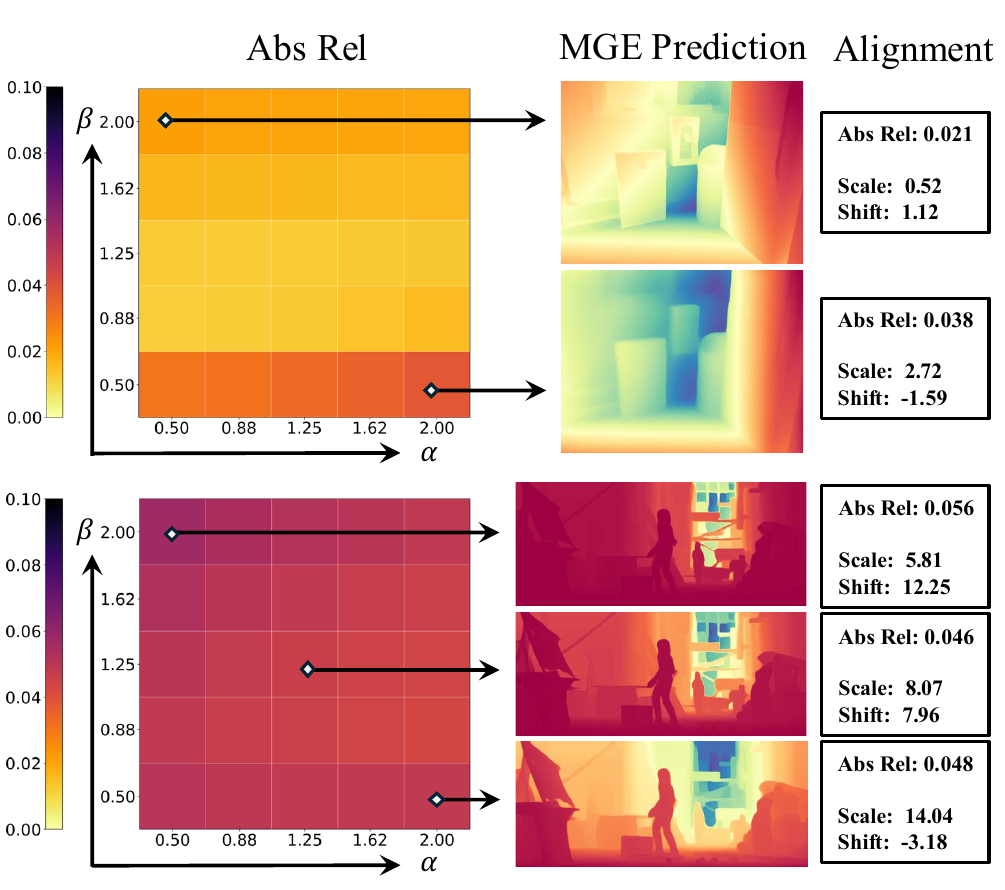}
  \vspace{-5pt}
  \caption{
  {Empirical study on scale-shift variations. 
  The left part illustrates the MGE performance (AbsRel) after modulating the latent features with sampled modulation parameters ($\alpha, \beta$). 
  The right part visualizes the corresponding MGE outputs and parameters used to align with GT. 
  It can be observed that despite the predicted depth exhibiting significant scale and shift fluctuations, the underlying geometric accuracy remains largely unchanged.}}
  \label{fig:empirical_study}
  \vspace{-15pt}
\end{figure}

In this paper, we argue that existing foundation MGE models are already well-equipped to be repurposed for streaming video depth estimation without retraining the entire network.
We posit that temporal inconsistency does not primarily arise from per-frame perceptual errors, but from inherent scale ambiguity: without a mechanism to maintain consistent scale and shift over time, 
% each frame is predicted in an independent, drifting coordinate space.
the model effectively predicts each frame in an independently drifting coordinate system.
% Notably, we observe that pretrained MGE models like MoGe~\cite{wang2025moge} already encode strong geometric understanding, when each frame’s prediction is \textit{individually aligned by a simple per-frame scale and shift}, the reconstructed 3D geometry becomes accurate and consistent (Fig.~\ref{fig:recon_evidence} (b)). 
Notably, pretrained models such as MoGe~\cite{wang2025moge} already encode strong geometric structure—when each frame is \textit{individually aligned with a simple per-frame scale and shift}, the reconstructed 3D geometry becomes accurate and temporally coherent (Fig.~\ref{fig:recon_evidence}b).
This implies that 
% temporal inconsistency mainly arises from \textit{frame-to-frame scale-shift instability}. 
temporal inconsistency is largely driven by \textit{frame-to-frame scale–shift instability} rather than deficient geometry.
% To further investigate this phenomenon, we conduct an empirical analysis examining how the global statistics of latent features affect the scale and shift of the predictions (Sec.~\ref{sec: scale and shift}). 
To further investigate this phenomenon, we conduct an empirical analysis of how global latent feature statistics influence predicted scale and shift (Sec.~\ref{sec: scale and shift}).
Our results reveal that \textit{scale/shift variations are tightly coupled with the mean and variance of latent features} extracted by the pretrained encoder (Fig.~\ref{fig:empirical_study}). 
This finding suggests that rather than retraining the entire model, temporal stability can be achieved by directly regulating these latent feature statistics of MGE models, specifically, their mean and variance.

Motivated by this observation, we propose \textbf{Dynamic Feature Normalization} (\textbf{DyFN}), a lightweight, learnable module that predicts and dynamically modulates the mean and variance of latent features over time to enforce consistent depth across frames, without compromising the fidelity or generalization of pretrained MGE models. 
% \textbf{DyFN} is a general framework applicable to both offline and online video settings. 
% For \textbf{offline video depth prediction}, \textbf{DyFN} employs a compact 3D convolutional block that jointly considers all frames to predict per-frame normalization parameters, achieving globally consistent geometry. 
For \textit{online streaming depth estimation}, DyFN incorporates a recurrent ConvGRU-based module that updates normalization parameters based on aggregated historical context. 
By finetuning only this \textbf{DyFN}, while keeping the pretrained encoder and decoder frozen, our method efficiently adapts existing MGE models to continuous inputs, achieving temporal coherence without sacrificing geometric accuracy. 
Extensive experiments across diverse benchmarks demonstrate that \textbf{DyFN} achieves state-of-the-art temporal stability in streaming scenarios (See Figure~\ref{fig: qualitative} and Table~\ref{tab:depth_methods_comparison}). 
It not only surpasses strong video-based methods, but also preserves the strong single-frame accuracy of pretrained MGE models. 
%Our findings highlight a scalable and data-efficient paradigm for continuous stream depth estimation, stabilizing pretrained monocular models via dynamic feature normalization rather than full model retraining.

In summary, our main contributions are threefold:
\begin{itemize}
\item We identify the principal source of temporal instability in pretrained monocular depth models: \textit{frame-to-frame scale--shift inconsistency} arising from fluctuations in latent feature statistics (mean and variance), 
thereby establishing a direct connection between feature distribution and temporal stability.
\item We propose \textbf{Dynamic Feature Normalization (DyFN)}, a lightweight and general stabilization module that dynamically modulates latent feature statistics to 
% enforce temporal consistency.
maintain consistent scale and shift over time.
\item We show that finetuning only this small stabilizer while freezing the pretrained MGE backbone achieves state-of-the-art temporal stability across diverse video benchmarks \textit{without degrading} single-frame accuracy or generalization, surpassing fully trained video-based models.
\end{itemize}

\section{Related Work}
\label{sec:related}
% Recent advances in monocular depth estimation span four major categories: relative depth estimation without real-world scale, metric depth estimation with absolute units, video depth estimation with a focus on temporal consistency, and multi-frame reconstruction methods leveraging geometric and temporal cues. 
% We review representative works in each direction and highlight how our streaming-based design differs from prior approaches.

\vspace{0.1in}\noindent\textbf{Relative Depth Estimation} 
Relative depth estimation has demonstrated robust generalization across diverse domains by predicting depth up to an unknown scale and shift~\cite{birkl2023midas, ranftl2020towards, MegaDepthLi18, ranftl2021vision, eftekhar2021omnidata, yang2024depth,yang2024depthv2}. 
Foundational works like MiDaS~\cite{ranftl2020towards} established this paradigm using multi-dataset training combined with scale-invariant losses. 
Subsequent research has increasingly integrated large-scale self-supervised pretraining~\cite{oquab2023dinov2, he2022masked, weinzaepfel2022croco, weinzaepfel2023croco, xie2017aggregated} to enhance feature representation. 
Notably, DPT~\cite{ranftl2021vision} successfully adapted transformers for dense prediction, a strategy further scaled by Depth Anything~\cite{yang2024depth,yang2024depthv2} utilizing over 60 million unlabeled images to achieve superior zero-shot performance. 
More recently, diffusion-based methods such as Marigold~\cite{ke2024repurposing} and GeoWizard~\cite{fu2024geowizard} have leveraged generative priors~\cite{rombach2022high, podell2023sdxl} to push the boundaries of detail recovery. 
Despite these advancements, these frame-centric approaches process images independently, inherently failing to maintain temporal consistency when applied to video streams.

\vspace{0.1in}\noindent\textbf{Metric Depth Estimation.}
To resolve scale-shift ambiguity, metric depth estimation aims to recover absolute depth from monocular inputs, a task that remains fundamentally ill-posed~\cite{yin2021learning, yin2023metric3d, Hu_2024, piccinelli2024unidepth, piccinelli2025unidepthv2universalmonocularmetric, wang2025moge2, guo2025depthcamerazeroshotmetric, bhat2023zoedepth}. 
Recent methods tackle this by incorporating strong geometric priors or optimizing for camera intrinsics. 
For instance, LeReS~\cite{yin2021learning} leverages scene statistics to align predictions, while ZoeDepth~\cite{bhat2023zoedepth} extends relative depth networks with adaptive metric bins to handle scene variability. Addressing the dependency on camera parameters, Metric3D~\cite{yin2023metric3d} and its successor~\cite{Hu_2024} propose zero-shot inference within a canonical camera space, whereas UniDepth~\cite{piccinelli2024unidepth} employs spherical parameterization to disentangle intrinsics for broader generalization. 
Although anchoring predictions to a metric scale theoretically reduces global scale drift, these methods process video frames individually. 
Without explicit temporal integration, they remain susceptible to inter-frame flickering and metric instability when applied to dynamic video streams.

\vspace{0.1in}\noindent\textbf{Video and Stream Depth Estimation}
To explicitly model temporal dependencies, recent works extend image-based baselines by injecting bidirectional attention~\cite{tan2023temporal}, incorporating recurrent networks~\cite{gu2024mamba, ballas2015delving, lstm, wang2025continuous}, or leveraging pre-trained video generative models~\cite{blattmann2023stable, wan2025wan}. For instance, Video Depth Anything~\cite{chen2025video} augments the static Depth Anything V2 architecture with spatial-temporal attention and keyframe scheduling to enhance long-term consistency. Similarly, RollingDepth~\cite{ke2024video} employs multi-frame cross-attention aligned with global optimization. In the generative domain, ChronoDepth~\cite{shao2024learningtemporallyconsistentvideo} pioneers the use of video diffusion priors for depth regression, while DepthCrafter~\cite{hu2024depthcrafter} adopts a curriculum-based training strategy to synthesize temporally coherent sequences. 
However, these window-based methods typically rely on processing fixed-length clips with overlapping inference. 
This paradigm inherently incurs high latency and memory redundancy, limiting their applicability for long-duration or real-time scenarios. 
Consequently, streaming architectures have emerged as an efficient alternative, processing arbitrary sequence lengths via recurrent states. 
A representative work, FlashDepth~\cite{chou2025flashdepth}, demonstrates this potential by maintaining a compact hidden state to achieve real-time inference at 2K resolution without sacrificing temporal stability.

\vspace{0.1in}\noindent\textbf{Multi-frame Geometry Estimation}
Distinct from direct depth regression, geometric estimation methods reconstruct dense 3D point maps directly from images, leveraging explicit geometric constraints to enhance structural fidelity~\cite{wang2024dust3r, wang2025continuous, cabon2025must3r, wang20243d, murai2025mast3r, zhang2024monst3r, wang2025vggt, yang2025fast3r, wang2025moge, wang2025moge2, chen2025ttt3r, sun2025unigeo}. 
The foundational work, Dust3R~\cite{wang2024dust3r}, reformulates pairwise structure-from-motion as a regression task, enabling robust reconstruction from uncalibrated views. 
This paradigm has been extended to dynamic and sequential contexts: MonST3R~\cite{zhang2024monst3r} incorporates motion priors to handle dynamic objects, while VGGT~\cite{wang2025vggt} utilizes global spatiotemporal transformers for consistent scene reconstruction. 
To address the memory bottlenecks in long-sequence processing, CUT3R~\cite{wang2025continuous} adopts LSTM-style recurrent updates to accumulate geometric features, and TTT3R~\cite{chen2025ttt3r} optimizes memory read-write mechanisms to enhance localization stability. 
However, these geometry-centric approaches generally incur significant computational overhead and heavily rely on multi-view overlap. 
Consequently, they often struggle in pure monocular settings or highly dynamic environments where consistent geometric constraints are violated or absent.
\section{Empirical Studies}
\label{sec: scale and shift}
Pretrained monocular geometry foundation models such as MoGe~\cite{wang2025moge} achieve remarkable accuracy on single images, reconstructing fine geometric details and demonstrating strong zero-shot generalization. 
Despite this single-frame success, their direct application to continuous video streams reveals a critical limitation. 
When applied naively in a frame-by-frame manner, these models suffer from severe scale-shift ambiguity across consecutive frames. 
Although each individual prediction may appear geometrically plausible in isolation, they are not anchored to a consistent 3D coordinate frame. 
This failure to maintain a stable geometric reference results in significant structural instability in the aggregated 3D scene. 
As illustrated in Fig.~\ref{fig:recon_evidence}c, this manifests as \textbf{non-rigid warping and geometric drift} over time, rather than a coherent, stable reconstruction. 
This discrepancy highlights a fundamental gap: while existing models encode powerful static spatial priors, they lack the cross-frame geometric coherence required for stable 3D reconstruction from continuous stream.

\subsection{What are the causes of temporal inconsistency?}  
To understand the root cause of this inconsistency, we perform an empirical study using the state-of-the-art monocular depth estimator MoGe~\cite{wang2025moge} as a representative model. 
We find that the model's underlying geometric understanding is already robust. 
When each frame’s prediction is individually aligned to the ground truth using a simple affine transformation (scale and shift), the reconstructed 3D geometry becomes highly accurate and geometric consistent (Fig.~\ref{fig:recon_evidence}b).
As demonstrated in Fig~\ref{fig:recon_evidence}c, when fusing the predicted point clouds using a single sequential alignment (one scale/shift for the entire sequence), the reconstruction suffers from severe non-rigid warping and geometric drift, achieving an accuracy ($\delta<1.25$) of only 62.5. 
In stark contrast, when we align each frame individually (per-frame scale and shift), the accuracy dramatically increases to 99.8.
This reveals that the primary cause of temporal inconsistency is not geometric degradation but rather \textit{frame-to-frame scale-shift variation}, the predicted global depth scale and offset drift over time, leading to unstable geometry estimation sequences.

\subsection{What are the causes of scale-shift variations?}   

We further investigate what causes such \emph{scale--shift fluctuations} in monocular geometry models. 
Most modern MGE networks share a common encoder--decoder architecture, where an input image $\mathcal{I}$ is first encoded into latent features $\mathcal{F} = \mathcal{E}(\mathcal{I})$, which are then decoded into a point map $\mathcal{P} = \mathcal{D}(\mathcal{F})$. 
%Given this structural similarity, we select the state-of-the-art \textbf{MoGe}~\cite{wang2025moge} as a representative foundation model for analysis.

\vspace{0.1in}\noindent\textbf{Analysis Setup.}
%We hypothesize that the \emph{latent feature distribution} plays a key role in determining the global scale and shift of the predicted depth. 
As the scale and shift parameters are global statistics that govern the predictions, we study how the global \emph{latent feature distribution} influences these parameters across frames. 
To examine this, we select two images from distinct domains (indoor and outdoor) and extract their latent features $\mathcal{F}$. 
For each feature map, we compute its channel-wise mean $\boldsymbol{{\mu}}_{\mathcal{F}}$ and standard deviation $\boldsymbol{\sigma}_{\mathcal{F}}$, then normalize the features as
\begin{equation}
\mathcal{F}_{\text{norm}} = \frac{\mathcal{F} - \boldsymbol{\mu}_{\mathcal{F}}}{\boldsymbol{\sigma}_{\mathcal{F}} + \epsilon},
\label{eq:normalize}
\end{equation}
where $\epsilon$ is a small constant for numerical stability.
We then introduce scaling multipliers $\alpha, \beta \in [0.5, 2.0]$ to modulate the mean and standard deviation, defining
\begin{equation}
\boldsymbol{\mu}_{\mathcal{F}}^{\alpha} = \alpha \cdot \boldsymbol{\mu}_{\mathcal{F}}, \qquad
\boldsymbol{\sigma}_{\mathcal{F}}^{\beta} = \beta \cdot \boldsymbol{\sigma}_{\mathcal{F}},
\end{equation}
and reconstructing modified features as
\begin{equation}
\mathcal{F}^{\alpha, \beta} = \mathcal{F}_{\text{norm}} \cdot \boldsymbol{\sigma}_{\mathcal{F}}^{\beta} + \boldsymbol{\mu}_{\mathcal{F}}^{\alpha}.
\label{eq:modulation}
\end{equation}
Each modified feature map $\mathcal{F}^{\alpha, \beta}$ is fed through the frozen decoder $\mathcal{D}$ to produce a new point map. 
Using least-squares fitting, we estimate the corresponding affine transformation (scale and shift) relative to the original output. 
The correlation between the modulation parameters $(\alpha, \beta)$ and the resulting geometric transformation is visualized in Fig.~\ref{fig:empirical_study}.

% As shown in Fig.~\ref{fig:empirical_study}, the baseline MoGe model produces accurate geometry for both indoor (ScanNet, AbsRel = 0.0116) and outdoor (Sintel, AbsRel = 0.0463) scenes. However, when we modulate $\alpha$ and $\beta$, the predictions' \emph{scale} and \emph{shift} vary substantially, while the relative geometric accuracy remains nearly unchanged.
% Specifically, the mean multiplier $\alpha$ exhibits an approximately \emph{linear} relationship with output scale and shift (solid lines), whereas the standard deviation multiplier $\beta$ shows a \emph{nonlinear} influence (dashed lines). For instance, in the outdoor scene, fixing $\beta = 1.21$ and varying $\alpha$ from $0.5$ to $2.0$ increases the output scale from $5.6$ to $10.0$ and alters the shift from $9.3$ to $7.8$, yet the AbsRel error changes only marginally ($0.050 \rightarrow 0.044$).
% we first sample a set of scaling multipliers, and apply them to the latent feature according to Eq.~\ref{eq:modulation}. 

% As shown in Fig.~\ref{fig:empirical_study}, our results show that while different modulation parameters lead to noticeable variations in the absolute scale and shift of the generated depth maps (ranging from [0.52, 12] and [-7.35, 6.98] respectively), the overall geometric structure remains consistent. 
% Moreover, after applying an post affine alignment to the ground-truth depth, the predictions still maintain high geometric accuracy. 
% We conduct experiments on both indoor (ScanNet) and outdoor (Sintel) datasets to verify the generality of this phenomenon.
\vspace{0.1in}\noindent\textbf{Empirical Results.}
As shown in Fig.~\ref{fig:empirical_study}, our experiments on both indoor (ScanNet) and outdoor (Sintel) datasets show a clear phenomenon. 
We found that altering the statistics of latent features (such as their mean and variance) directly causes the absolute scale and shift of the predicted depth maps to change dramatically (ranging from [0.52, 14.04] and [-3.18, 12.25], respectively). 
However, even when the scale and shift changed this much, the actual geometric shape of the prediction remained surprisingly stable. 
We verified this by applying the affine alignment to match the predictions with the ground truth; after alignment, the geometric accuracy was still very high.
These results confirm our central hypothesis: \emph{the mean and variance of latent features are strongly coupled with the prediction's scale and shift}, but are \emph{largely decoupled from its relative geometric accuracy}. 
In essence, uncontrolled fluctuations in these feature statistics across a video stream are the direct cause of the observed scale-shift drift, which in turn manifests as the structural inconsistency detailed previously. 
This key insight directly motivates our proposed \textit{Dynamic Feature Normalization}, a lightweight mechanism designed to explicitly regulate these statistics over time to enforce stable and geometrically coherent point predictions, as detailed in the following section.

\section{Dynamic Feature Normalization}

\begin{figure*}[thbp]
  \vspace{-5pt}
  \centering
  \includegraphics[width=0.9\textwidth]{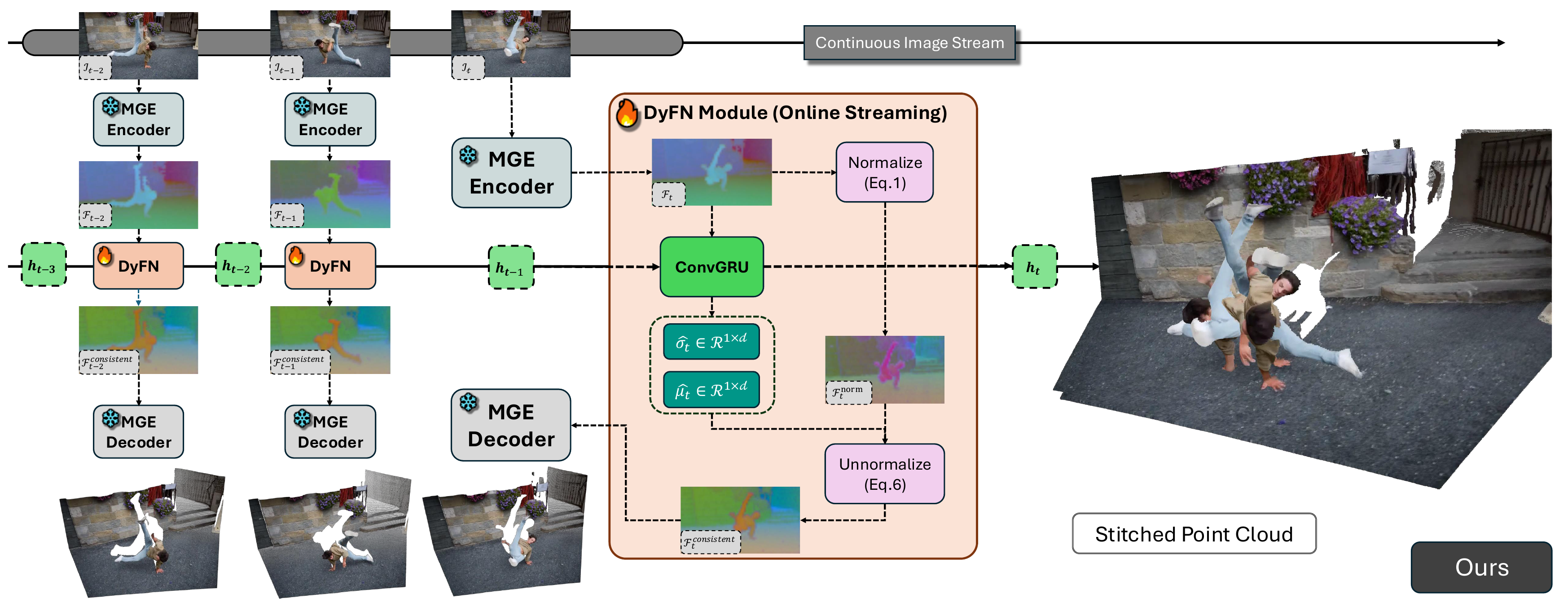}
  \caption{\textbf{Method Overview.} Our method performs consistent geometry estimation from an image stream. Each frame is processed by a shared ViT-based MGE encoder to extract visual features $\mathcal{F}_t$. These features are then passed into our recurrent \textit{Dynamic Feature Normalization (DyFN)} module. The DyFN module leverages a temporally-aware hidden state $\boldsymbol{h}$ to dynamically modulate the mean and variance of $\mathcal{F}_t$, producing temporally consistent features $\mathcal{F}_t^{consistent}$. These stabilized features are subsequently fed into the MGE decoder to regress a consistent point map. Finally, a correspondence-based rigid pose solver (estimating rotation and translation) aggregates these point maps to produce a stable and coherent 3D reconstruction (See supplementary for more details).}
  \label{fig:pipeline_placeholder}
  \vspace{-12pt}
\end{figure*}

Inspired by the empirical findings in Sec.~\ref{sec: scale and shift}, we propose the {Dynamic Feature Normalization (DyFN)} module to address the scale-shift inconsistency inherent to pretrained monocular geometry model. 
DyFN dynamically modulates latent features based on their temporal context to enforce stable and consistent geometry predictions for \textit{online streaming video}. 
As illustrated in Fig.~\ref{fig:pipeline_placeholder}, our approach freezes the pretrained encoder and decoder. 
Only the lightweight DyFN module is trained, allowing it to adapt the feature statistics for temporal consistency.

This module first normalizes the incoming latent feature $\mathcal{F}_t$ to obtain a standardized feature $\mathcal{F}^{\text{norm}}_t$ following the Eq.~\ref{eq:normalize}.
This feature is fed into a {Convolutional GRU (ConvGRU)}~\cite{ballas2015delving}, which maintains a hidden state $\boldsymbol{h}_t$ that summarizes observations from all previous frames. 
At each timestep $t$, the ConvGRU updates its state based on the previous state $\boldsymbol{h}_{t-1}$ and the current feature $\mathcal{F}_t$:
\begin{equation}
\boldsymbol{h}_t = \text{ConvGRU}(\mathcal{F}_t, \boldsymbol{h}_{t-1}).
\label{eq:convgru}
\end{equation}
For the first frame ($t=1$), the hidden state $\boldsymbol{h}_0$ is initialized as a zero tensor.

Two lightweight $1 \times 1$ convolutional heads then project the hidden state $\boldsymbol{h}_t$ to predict the spatial modulation parameters, a mean $\hat{\boldsymbol{\mu}}_t$ and a standard deviation $\hat{\boldsymbol{\sigma}}_t$:
\begin{equation}
\hat{\boldsymbol{\sigma}}_t = \text{Conv}_{1 \times 1}^{{\sigma}}(\boldsymbol{h}_t), \quad \hat{\boldsymbol{\mu}}_t = \text{Conv}_{1 \times 1}^{\mu}(\boldsymbol{h}_t).
\label{eq:heads}
\end{equation}
These predicted statistics are used to modulate the normalized feature, effectively replacing the original, unstable per-frame statistics with temporally-aware ones:
\begin{equation}
\mathcal{F}_t^{\text{consistent}} = \hat{\boldsymbol{\sigma}}_t \cdot \mathcal{F}_t^{\text{norm}} + \hat{\boldsymbol{\mu}}_t.
\label{eq:renorm_general}
\end{equation}
This final re-modulated feature $\mathcal{F}_t^{\text{consistent}}$ is then passed to the frozen decoder $\mathcal{D}$ to predict the final depth map $P_t$.

\section{Model Training} 
\label{sec: Model Training} 
%\subsection{Model Training}
%\label{subsec:details}
\vspace{0.1in}\noindent\textbf{Trained Modules.}
Guided by our analysis that the pretrained monocular geometry model already captures robust \emph{relative} geometric knowledge, we adopt a parameter-efficient fine-tuning strategy. 
We freeze the weights of both the encoder $\mathcal{E}$ and decoder $\mathcal{D}$ to preserve their powerful, generalized feature representations. 
Temporal consistency is then achieved by training \textbf{only} the lightweight DyFN module, which is tasked with modulating the latent feature distributions. 
This approach is highly efficient, as the DyFN module constitutes merely $2\%$ of the total parameters. 
As demonstrated in Tab.~\ref{tab:depth_methods_comparison} and Tab.~\ref{tab: image_depth}, this strategy successfully achieves our dual objectives: it retains the strong per-frame geometric accuracy of the frozen backbone while efficiently enforcing sequence-level stability.

\vspace{0.1in}\noindent\textbf{Training Objective.} 
In addition to the base loss from the backbone, $\mathcal{L}_{\text{MoGe}}$, we introduce two terms that explicitly supervise scale-shift stability. 
The first is a \emph{global alignment loss}, $\mathcal{L}_{\text{align}}$, designed to {enforce a single, consistent scale and shift across the entire sequence.} 
Given a sequence of $L$ predicted point maps $\{\hat{\rm P}_j\}_{j=1}^{L}$, we compute a \textit{single} global affine pair $(s_g, t_g)$ that best aligns all points from all frames to the ground truth simultaneously (e.g., via least-squares). 
The loss is then defined as the total error measured \textit{only} against this single global transformation:
\begin{equation}
\label{eq:global_align}
\mathcal{L}_{\text{align}}
= \sum_{j=1}^{L} \sum_{i \in \mathcal{M}} \frac{1}{z_i}
\big\lVert s_g \hat{\rm p}^{\,i}_j + t_g - \rm p^{\,i}_j \big\rVert_{1},
\end{equation}
where $\rm p^{\,i}_j$ is the $i$-th ground-truth point in frame $j$ with depth $z_i$, $\hat{\rm p}^{\,i}_j$ is the corresponding prediction, and $\mathcal{M}$ denotes valid points. This formulation directly penalizes any frame $j$ whose prediction $\hat{\rm p}_j$ deviates from the sequence-wide optimal scale $s_g$ and shift $t_g$. 
This forces the network's underlying feature representation (e.g., via DyFN) to produce outputs that are inherently stable over time.

The second term is an \emph{inter-frame temporal loss}, $\mathcal{L}_{\text{temp}}$, designed to mitigate long-horizon drift. 
It enforces that the \textbf{magnitude of temporal change} in the predictions matches that of the ground truth. To capture both short- and long-term dynamics, we compute this loss over multiple window sizes $k \in K=\{1,2,4\}$. The loss penalizes the scaled L1-discrepancy between the predicted and ground-truth inter-frame deltas:
\begin{equation}
\label{eq:temp_loss}
\mathcal{L}_{\text{temp}}
= \sum_{k \in K} \sum_{j=1}^{L-k} \sum_{i \in \mathcal{M}} \frac{1}{z_i}
\left\lVert
\, s_g \hat{\delta}^{\,i}_{j,k} - \delta^{\,i}_{j,k} \right\rVert_{1},
\end{equation}
where $\hat{\delta}^{\,i}_{j,k} = \big\lVert \hat{\rm p}^{\,i}_j - \hat{\rm p}^{\,i}_{j+k} \big\rVert_{1}$ represents the magnitude of the predicted change for point $i$ across $k$ frames, and $\delta^{\,i}_{j,k} = \big\lVert \rm p^{\,i}_j - \rm p^{\,i}_{j+k} \big\rVert_{1}$ is the corresponding ground-truth change. 
Applying the global scale $s_g$ (derived from $\mathcal{L}_{\text{align}}$) ensures that the predicted deltas are measured in the same metric space as the ground-truth deltas before comparison.

Our final objective is a weighted sum:
\begin{equation}
\mathcal{L}_{\text{final}} 
= \mathcal{L}_{\text{MoGe}} 
+ \alpha\, \mathcal{L}_{\text{align}} 
+ \beta\, \mathcal{L}_{\text{temp}},
\end{equation}
where $\alpha=1$ and $\beta=0.1$ in our training. 
Details of $\mathcal{L}_{\text{MoGe}}$ are provided in the supplementary material.

\vspace{0.1in}\noindent \textbf{Training Data.}
To ensure robustness across diverse scenarios, we finetune only the DyFN module on a large-scale data compilation. 
% This training mixture integrates seven distinct datasets: \emph{Dynamic Replica}~\cite{karaev2023dynamicstereo}, \emph{IRS}~\cite{wang2021irslargenaturalisticindoor}, \emph{KenBurns3D}~\cite{Niklaus_TOG_2019}, \emph{Spring}~\cite{Mehl2023_Spring}, \emph{TartanAir}~\cite{tartanair2020iros}, \emph{MidAir}~\cite{Fonder2019MidAir}, and \emph{PointOdyssey}~\cite{zheng2023point}. 
This combined corpus provides approximately 1M total frames for training. 
Unless otherwise noted, our training procedure involves sampling fixed-length, continuous clips of 12 frames. 
More details are shown in the supplementary material.

\section{Experiment}
\label{sec:experiment}

\paragraph{Baselines} 
To comprehensively evaluate our method on video depth estimation, we compare it against representative works from six distinct paradigms. 
We broadly categorize these approaches based on their output: \emph{Depth Estimation} methods produce only per-frame depth, while \emph{Geometry} methods output per-frame point maps.
The specific categories are as follows:
(1)~\emph{Relative Depth Estimation:} Single-image models trained with an affine-invariant loss. 
As they process frames independently, they lack temporal consistency (e.g., Marigold~\cite{ke2024repurposing}, Depth Anything V1\&V2 (``DAV1, DAV2'')~\cite{yang2024depth, yang2024depthv2}, MoGe~\cite{wang2025moge}).
(2)~\emph{Metric Depth Estimation:} Single-image models that are trained to predict depth at a true metric scale (e.g., DepthPro~\cite{bochkovskiydepth}, MoGe v2~\cite{wang2025moge2}).
(3)~\emph{Multi-frame Geometry:} Offline models that process a set of views and leverage mechanisms like bidirectional cross-attention to enforce multi-view consistency (e.g., VGGT~\cite{wang2025vggt}, Monst3R~\cite{zhang2024monst3r}).
(4)~\emph{Streaming Geometry:} Methods that support online inputs, fusing temporal information using recurrent modules(e.g., CUT3R~\cite{wang2025continuous}, TTT3R~\cite{chen2025ttt3r}).
(5)~\emph{Video Depth Estimation:} Models that use temporal fusion but are constrained to fixed-length, offline video inputs (e.g., DepthCrafter~\cite{hu2024depthcrafter}, VideoDepthAnything(``VDA'')~\cite{shao2024learningtemporallyconsistentvideo}).
(6)~\emph{Streaming Depth Estimation:} Methods support online streaming inputs and use recurrent modules to update features for temporal consistency (e.g., FlashDepth~\cite{chou2025flashdepth}).

\vspace{0.1in}\noindent\textbf{Datasets and Metrics.}
For quantitative evaluation, we follow the protocol of DepthCrafter~\cite{hu2024depthcrafter} and select representative scenes from datasets covering indoor~\cite{dai2017scannet, palazzolo2019iros}, outdoor~\cite{kitti}, and in-the-wild environments~\cite{Butler:ECCV:2012}.
We evaluate geometric accuracy using the Absolute Relative Error (AbsRel) and the $\delta<1.25$ threshold. 
Detailed formulations for these metrics are available in the supplementary material.

\vspace{0.1in}\noindent\textbf{Evaluation Protocol.}
For non-metric models, we first align their predictions to the ground-truth scale using a least-squares method. A crucial distinction lies in the {scope} of this alignment, which we define for two separate evaluation settings:
\textbf{(1) Video Depth Evaluation:} We compute a {single} scale and shift for the {entire sequence} and apply it uniformly to all frames. This protocol stringently evaluates both per-frame accuracy and, critically, inter-frame temporal consistency.
\textbf{(2) Image Depth Evaluation:} We compute a {separate} scale and shift for {each frame independently}. This protocol isolates per-frame accuracy, measuring the model's static performance without penalizing temporal instability.

\begin{table}[thbp]
  \centering
    \setlength{\tabcolsep}{2pt}
    % \resizebox{\linewidth}{!}{
    \adjustbox{width={\linewidth},keepaspectratio}{
  \begin{tabular}{@{}lcccccccc@{}}
    \toprule
    \multirow{2}{*}{} & \multicolumn{2}{c}{\textbf{Sintel (50 frames)}} & \multicolumn{2}{c}{\textbf{Scannet (90 frames)}} & \multicolumn{2}{c}{\textbf{KITTI (110 frames)}} & \multicolumn{2}{c}{\textbf{Bonn (110 frames)}} \\
    \cmidrule(lr){2-3} \cmidrule(lr){4-5} \cmidrule(lr){6-7} \cmidrule(lr){8-9}
     \textbf{Method} & Abs Rel$\downarrow$ & $\delta < 1.25 \uparrow$ & Abs Rel$\downarrow$ & $\delta < 1.25 \uparrow$ & Abs Rel$\downarrow$ & $\delta < 1.25 \uparrow$ & Abs Rel$\downarrow$ & $\delta < 1.25 \uparrow$ \\
    \midrule
    % \catA~Marigold            & 0.307 & 57.0 & 0.144 & 78.8 & 0.126 & 85.0 & 0.077 & 94.9 \\
    \catA~Marigold            & 0.532 & 51.5 & 0.166 & 76.9 & 0.149 & 79.6 & 0.091 & 93.1 \\
    \catA~DAV1   & 0.325 & 56.4 & 0.130 & 83.8 & 0.142 & 80.3 & 0.078 & 93.9 \\
    \catA~DAV2   & 0.367 & 55.4 & 0.135 & 82.2 & 0.140 & 80.4 & 0.106 & 92.1 \\
    \catA~MoGe v1             & 0.216 & 65.3 & 0.117 & 84.7 & 0.076 & 96.0 & 0.074 & 95.5 \\
    \specialrule{1pt}{1pt}{1pt} % Thick line separator
    % \catB~Metric3Dv2             & 0.213 & 68.6 & 0.063 & 96.6 & 0.055 & 98.0 & 0.067 & 97.1 \\
    % \catB~DepthPro             & 0.319 & 52.0 & 0.088$^*$ & 92.7$^*$ & 0.088$^*$ & 92.2$^*$ & 0.063$^*$ & 96.6$^*$ \\
    % \catB~MoGE v2              & 0.214 & 69.5 & 0.110$^*$ & 88.2$^*$ & 0.183$^*$ & 58.8$^*$ & 0.104$^*$ & 91.2$^*$ \\
    \catB~DepthPro             & 0.319 & 52.0 & (0.088) & (92.7) & (0.088) & (92.2) & (0.063) & (96.6) \\
    \catB~MoGe v2              & 0.214 & 69.5 & (0.110) & (88.2) & (0.183) & (58.8) & (0.049) & (98.0) \\
    \specialrule{1pt}{1pt}{1pt} % Thick line separator
    \catC~VGGT                & 0.287 & 66.1 & \color{gray}{0.031} & \color{gray}{98.5} & 0.070 & 96.5 & 0.055 & 97.1 \\
    \catC~Monst3R              & 0.335 & 58.5 & 0.123 & 83.2 & 0.104 & 89.5 & 0.063 & 96.4 \\
    \specialrule{1pt}{1pt}{1pt} % Thick line separator
    \catD~CUT3R               & 0.421 & 47.9 & \color{gray}{0.097} & \color{gray}{88.7} & 0.118 & 88.1 & 0.078 & 93.7 \\
    \catD~TTT3R               & 0.404 & 50.0 & \color{gray}{0.114} & \color{gray}{87.7} & 0.113 & 90.4 & 0.068 & 95.4 \\
    % \catD~StreamingVGGT        & 0.323 & 65.7 & - & - & 0.173  & 72.1 & 0.059 & 97.2 \\
    \specialrule{1pt}{1pt}{1pt} % Thick line separator
    \catE~DepthCrafter        & 0.270 & 69.7 & 0.123 & 85.6 & 0.104 & 89.6 & 0.071 & 97.2 \\
    \catE~VDA  & 0.300 & 63.3 & 0.075 & 95.4 & 0.079 & 95.0 & 0.051 & 98.1 \\
    % \catE~Rolling Depth       & - & - & 0.093 & 91.6 & - & - & 0.079 & 93.9 \\
    \specialrule{1pt}{1pt}{1pt} % Thick line separator
    \catF~FlashDepth          & 0.265 & 64.2 & 0.101 & 90.3 & 0.103 & 89.5 & 0.053 & 98.0 \\
    \catF~Ours                & \textbf{0.180} & \textbf{73.0} & \textbf{0.073} & \textbf{96.6} & \textbf{0.062} & \textbf{97.3} & \textbf{0.044} & \textbf{98.4} \\
    \bottomrule
  \end{tabular}%
  } % Closing brace for \resizebox
    \caption{\textbf{Quantitative evaluation of video depth estimation} on the Sintel, ScanNet, KITTI, and Bonn datasets.
    We compare methods across six categories:
    \catA{} Relative Depth,
    \catB{} Metric Depth,
    \catC{} Multi-frame Geometry,
    \catD{} Streaming Geometry,
    \catE{} Video Depth, and
    \catF{} Streaming Depth.
    The best results are highlighted in \textbf{bold}.
    Values in \textcolor{gray}{gray} indicate that the method was trained on the target dataset.
    Values in parentheses denote evaluations performed on the raw metric output without alignment.
    }
  \label{tab:depth_methods_comparison}
    \vspace{-12pt}
\end{table}

\subsection{Monocular and Video Depth Estimation}
\label{subsec: monocular and video}

\begin{figure}[htbp]
  \centering
  \vspace{-10pt}
  \includegraphics[width=\linewidth]{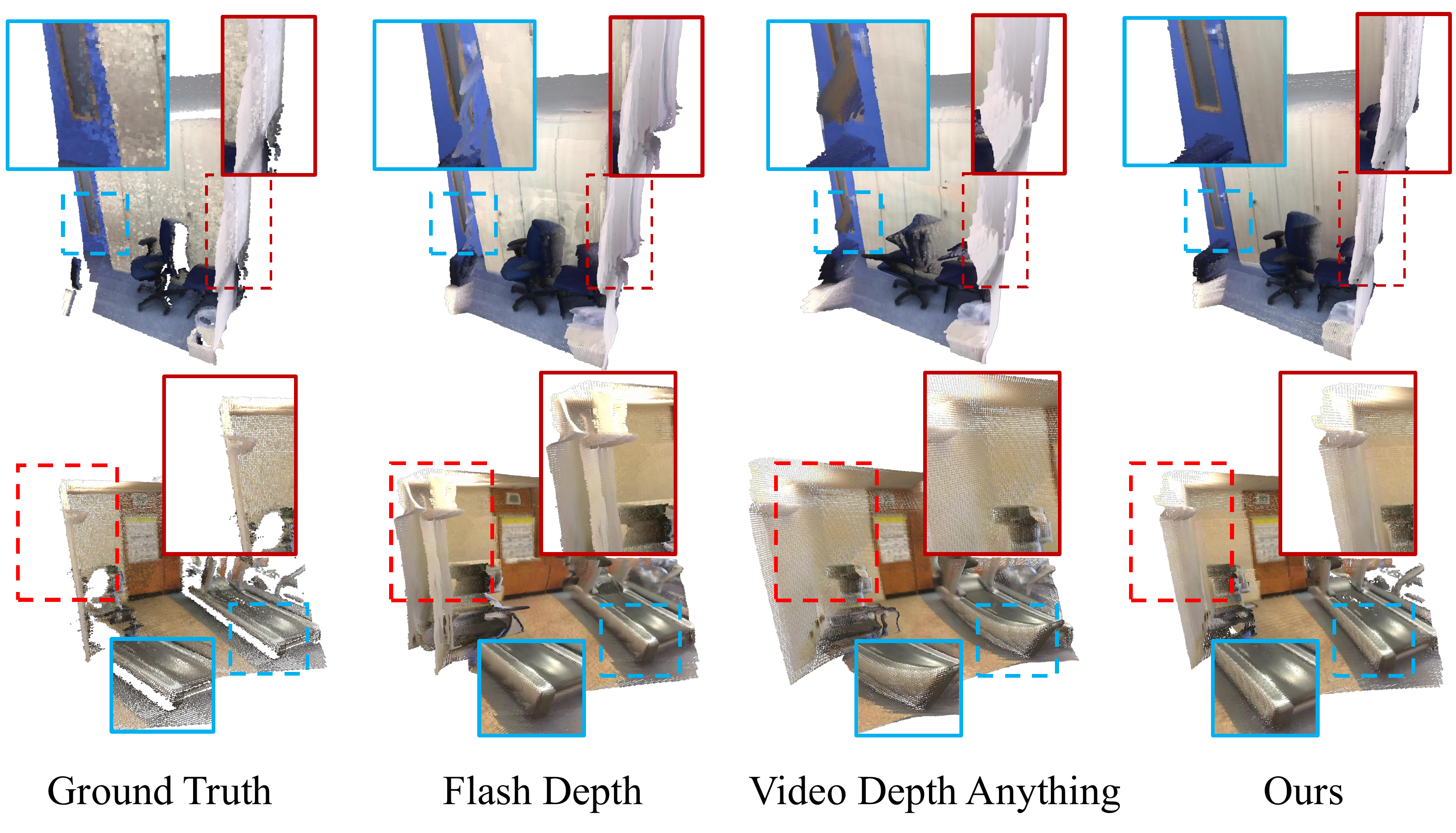}
  \vspace{-20pt}
  \caption{Qualitative comparison on indoor scenes. We compare our method with Flash Depth and VDA. Our method shows the best geometric consistent and less non-rigid warping. }
  \label{fig: qualitative}
   \vspace{-12pt}
\end{figure}

\paragraph{Video Depth Estimation}
As shown in Table~\ref{tab:depth_methods_comparison}, we conduct a comprehensive quantitative comparison of our proposed method against six distinct categories of existing works across four benchmarks (Sintel, ScanNet, KITTI, and Bonn). 
The results clearly demonstrate that our method achieves state-of-the-art performance, outperforming all other models across all datasets and metrics (AbsRel $\downarrow$ and $\delta < 1.25 \uparrow$).
(1)~Comparison with Monocular Depth Models (Categories \catA{} and \catB{}):
Single-image models suffer from a lack of temporal constraints. 
Their inherent scale-shift inconsistency is heavily penalized by the video evaluation protocol, which uses a single scale and shift for the entire sequence. 
Our method, by explicitly enforcing temporal consistency, shows significant improvements; for example, on Scannet, our $\delta < 1.25$ (96.6) is a \textbf{11.9\%} improvement over MoGe v1 (84.7).
Metric depth models, while trained to align with a metric scale, are similarly unstable and lack temporal fusion. 
This leads to volatile performance across datasets. 
For instance, MoGe v2's AbsRel on KITTI is 0.183, drastically worse than our 0.062. 
This highlights the superior stability and consistency of our approach.
(2)~Comparison with Multi-view Geometry Estimator (Categories \catC{} and \catD{}):
Multi-view geometry estimation methods mainly rely on static scene assumptions and known poses. 
Consequently, they perform poorly in dynamic scenes like Sintel~(e.g., VGGT AbsRel 0.287 vs. our 0.180 on Sintel).
Notably, while VGGT performs well on the static ScanNet dataset (0.031), this result is attributable to it being trained on this specific dataset (indicated by \textcolor{gray}{gray} text) and is not indicative of generalizability. 
On the static Bonn benchmark, our method (0.044) still surpasses VGGT (0.055) on Abs Rel. 
Furthermore, streaming-based reconstruction methods show an even more significant performance drop, confirming the limitations of their temporal fusion mechanisms.
(3)~Comparison with Video/Streaming Depth Models (Categories \catE{} and \catF{}):
As our empirical study suggests, our method addresses the root cause of temporal inconsistency through dynamic feature normalization. 
As a result, our performance is not only significantly better than other streaming-based methods like FlashDepth (e.g., 96.6 vs. 90.3 AbsRel on Scannet), but it also surpasses offline video depth methods that utilize more complex bidirectional attention mechanisms.
The qualitative results in Fig.~\ref{fig: qualitative} provide visual confirmation. 
For this visualization, we align predictions to the metric scale and transform point clouds into the global coordinate system using ground-truth poses. 
Our method's reconstructions exhibit superior geometric coherence and markedly less non-rigid warping compared to both FlashDepth and Video Depth Anything. 
Consequently, our method sets a new state-of-the-art, demonstrating that regulating feature statistics, as motivated by our empirical study (Sec.~\ref{sec: scale and shift}), is a highly effective strategy for achieving robust 3D consistency from streaming input.

\begin{table}[tbp]
    \centering 
    \setlength{\tabcolsep}{2pt}
    % \resizebox{\linewidth}{!}{
    \adjustbox{width={\linewidth},keepaspectratio}{
    \begin{tabular}{lcccccccc}
        \bottomrule
        \multirow{2}{*}{} & \multicolumn{2}{c}{\textbf{Sintel }} & \multicolumn{2}{c}{\textbf{Scannet}} & \multicolumn{2}{c}{\textbf{KITTI}} & \multicolumn{2}{c}{\textbf{Bonn}} \\
    \cmidrule(lr){2-3} \cmidrule(lr){4-5} \cmidrule(lr){6-7} \cmidrule(lr){8-9}
     \textbf{Method} & Abs Rel$\downarrow$ & $\delta < 1.25 \uparrow$ & Abs Rel$\downarrow$ & $\delta < 1.25 \uparrow$ & Abs Rel$\downarrow$ & $\delta < 1.25 \uparrow$ & Abs Rel$\downarrow$ & $\delta < 1.25 \uparrow$ \\
        \midrule
        \catA{} DAV2        & 0.200  & 74.1 & 0.039  & 98.2  & 0.073 & 95.3 & 0.048 & 98.0 \\
        \catA{} MoGe v1 & \textbf{0.124}  & \textbf{83.7} & \textbf{0.027}  & \textbf{98.6} & \textbf{0.044} & \textbf{98.0} & \textbf{0.028} & \textbf{98.8} \\
        \catD{} CUT3R & 0.428  & 55.4 & 0.064  & 93.7 & 0.092 & 91.3 & 0.063 & 96.2 \\
        \catE{} VDA & 0.200  & 75.3 & 0.041  & 98.1 & 0.074 & 95.1 & 0.039 & 98.6 \\
        \catF{} FlashDepth & 0.174  & 75.6 & 0.056  & 96.3 & 0.085 & 92.6 & 0.043 & 98.7 \\
        % \catF{} Ours(with DyFN)        & 0.134  & 82.2 & 0.028  & \textbf{98.7} & 0.047 & 97.7 & 0.029 & \textbf{98.8} \\
        \catF{} Ours    & \textbf{0.124}  & \textbf{83.7} & \textbf{0.027}  & \textbf{98.6} & \textbf{0.044} & \textbf{98.0} & \textbf{0.028} & \textbf{98.8} \\
        \toprule
    \end{tabular}}
    \vspace{-8pt}
    \caption{Single-frame depth evaluation. We report performance across four different categories: \catA{} Relative Depth, \catD{} Streaming Geometry, \catE{} Video Depth, and \catF{} Streaming Depth. Evaluations are performed on the Sintel, ScanNet, KITTI, and Bonn datasets. All models accept only a single image as input at a time.}
    \label{tab: image_depth}
    \vspace{-20pt}
\end{table}

\paragraph{Single Frame Depth Estimation.}
In addition to video-based metrics, we conduct a single-frame depth evaluation, with results shown in Table~\ref{tab: image_depth}. 
A key advantage of our methodology is that by freezing the pretrained encoder and decoder, our model \textbf{perfectly inherits the per-frame accuracy of its base model (MoGe v1)}. 
As the table demonstrates, our results are identical to MoGe v1 across all four datasets.
This is a critical distinction from other finetuning approaches, which often suffer from accuracy degradation. 
For example, FlashDepth (which builds on DepthAnything v2) sees its $\delta < 1.25$ score on KITTI drop from 95.3 (the base model's score) to 92.6 after finetuning. 
Our method avoids this tradeoff entirely. 
By preserving the base model's state-of-the-art accuracy, our approach achieves the best results across all datasets when compared to all other video and streaming-based methods.

\subsection{Ablation Study}
We conduct comprehensive ablation studies to validate our design choices, focusing on four key aspects: the effectiveness of the DyFN module, the contribution of loss functions, the choice of recurrent unit, and the global alignment strategy. 
Quantitative results are reported in Table~\ref{tab: ablation}.
All ablated experiments are trained on the full dataset as described on Sec.~\ref{sec: Model Training}, more details can be seen in the supplementary.

\noindent \textbf{Effectiveness of DyFN.} 
We demonstrate the impact of our proposed DyFN module. 
Compared to the MoGe, the introduction of DyFN yields significant improvements across all benchmarks. 
This confirms that the module effectively enhances temporal accuracy in video depth estimation.

\noindent \textbf{Impact of Loss Functions.} 
The removal of the alignment loss results in a sharp performance drop, even degrading accuracy beyond the baseline. 
This indicates that global-level alignment supervision is critical; without it, the DyFN module fails to learn the correct scale needed for alignment, negatively affecting relative depth precision. 
Additionally, the temporal loss ($\mathcal{L}_\text{temp}$) further refines the results by enforcing slight improvements in temporal consistency.

\noindent \textbf{Recurrent Unit Selection.} 
We compare different recurrent structures within DyFN. 
While both GRU and ConvGRU improve temporal consistency, the ConvGRU variant achieves superior performance. 
We attribute this to ConvGRU's ability to better capture spatially structured temporal statistics (i.e., stable mean and variance) compared to the standard GRU. 
See Supplementary for detailed results.

\noindent \textbf{Global Scale Alignment Strategy.} 
Finally, we evaluate the calculation method for the global scale $s_g$ and shift $t_g$ used in $\mathcal{L}_{align}$. 
We compare ``First Frame Alignment'' (denoted by $^\dagger$, aligning based on the first frame's prediction and GT) against ``Global Alignment'' (denoted by $^\ddagger$, derived from the entire sequence). 
Results show that the First Frame strategy outperforms the Global strategy. 
We observe that using the full sequence for alignment complicates optimization, as initial predictions are unstable, and significantly increases training overhead. 
Consequently, we adopt the First Frame Alignment strategy for our final model.

\begin{table}[t]
    \centering 
    % \resizebox{\linewidth}{!}{
    \setlength{\tabcolsep}{2pt}
    \adjustbox{width={\linewidth},keepaspectratio}{
    \begin{tabular}{lcccccccc}
        \bottomrule
        \multirow{2}{*}{} & \multicolumn{2}{c}{\textbf{Sintel }} & \multicolumn{2}{c}{\textbf{Scannet}} & \multicolumn{2}{c}{\textbf{KITTI}} & \multicolumn{2}{c}{\textbf{Bonn}} \\
    \cmidrule(lr){2-3} \cmidrule(lr){4-5} \cmidrule(lr){6-7} \cmidrule(lr){8-9}
     \textbf{Method} & Abs Rel$\downarrow$ & $\delta < 1.25 \uparrow$ & Abs Rel$\downarrow$ & $\delta < 1.25 \uparrow$ & Abs Rel$\downarrow$ & $\delta < 1.25 \uparrow$ & Abs Rel$\downarrow$ & $\delta < 1.25 \uparrow$ \\
        \midrule
        \catA{}MoGe        & 0.216 & 65.3 & 0.117 & 84.7 & 0.076 & 96.0 & 0.074 & 95.5 \\
        \catA{}Ours$^\dagger$        & 0.180 & 73.0 & 0.073 & 96.6 & 0.062 & 97.3 & 0.044 & 98.4 \\
        \midrule
        \catB{}w/o $\mathcal{L}_{align}$  & 0.245  & 61.8 & 0.124 & 83.1 & 0.088 & 93.5 & 0.088 & 93.3 \\
        \catB{}w/o $\mathcal{L}_{temp}$  & 0.183  & 72.7 & 0.069 & 96.4 & 0.063 & 97.0 & 0.044 & 98.4 \\
        \midrule
        \catC{}DyFN (convgru) $^\dagger$ & 0.180 & 73.0 & 0.073 & 96.6 & 0.062 & 97.3 & 0.044 & 98.4 \\
        \catC{}DyFN (gru) $^\dagger$ & 0.187  & 72.5 & 0.078 & 94.9 & 0.065 & 96.8 & 0.053 & 98.1 \\
        \midrule
        \catD{}$\mathcal{L}_{align}^\dagger$ & 0.180 & 73.0 & 0.073 & 96.6 & 0.062 & 97.3 & 0.044 & 98.4 \\
        \catD{}$\mathcal{L}_{align}^\ddagger$ & 0.189 & 72.1 & 0.066 & 96.4 & 0.070 & 96.2 & 0.045 & 98.3 \\
        \toprule
    \end{tabular}}
    \caption{Ablation studies. We conduct four different ablation studies, including: \catA{} Effectiveness of DyFN, \catB{} Impact of loss function, \catC{} Recurrent Unit Selection and \catD{} Global Scale Alignment Strategy. $^\dagger$ means that the model was trained with first frame alignment strategy. $^\ddagger$ means that the model was trained with the global frame alignment strategy.}
    \label{tab: ablation}
    \vspace{-12pt}
\end{table}

\subsection{Long Sequence Performance}
\begin{figure}[hbp]
  \vspace{-10pt}
  \centering
  \includegraphics[width=0.95\linewidth]{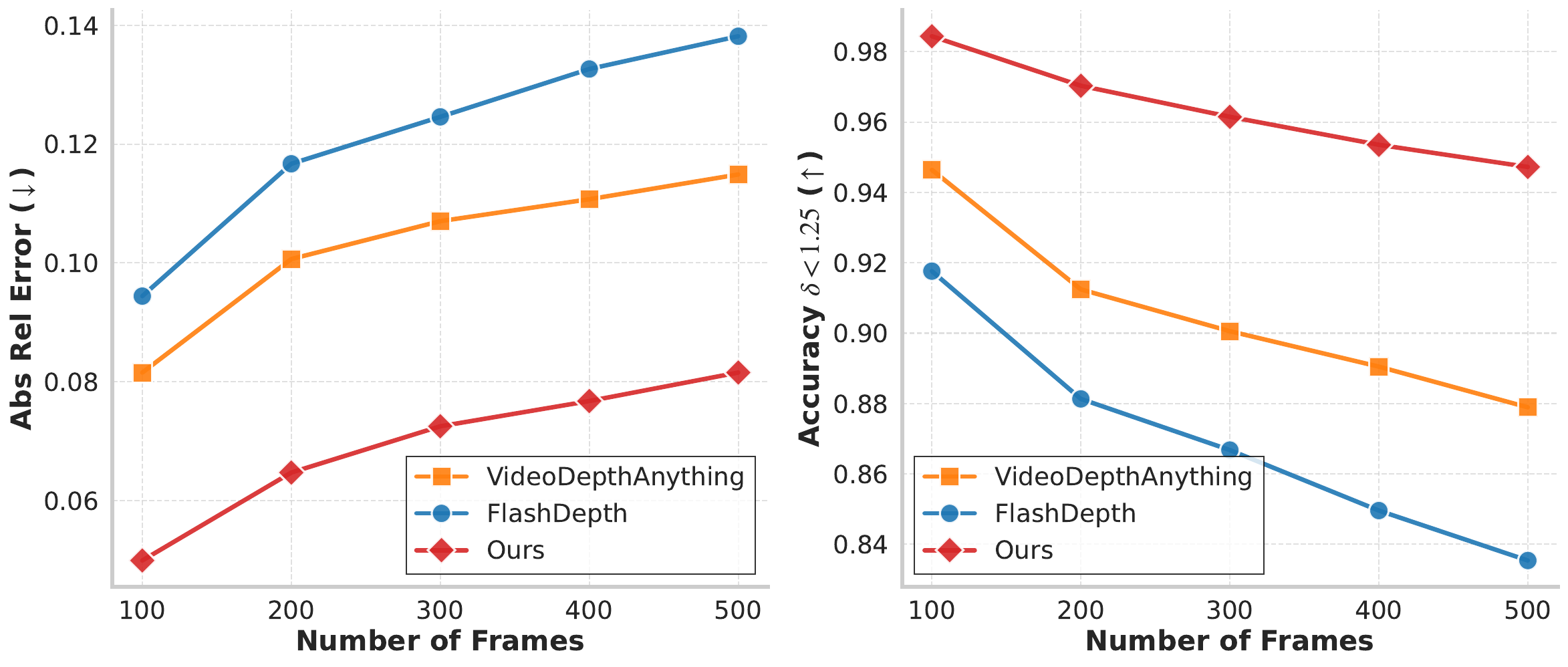}
  \vspace{-10pt}
  \caption{\textbf{Long-sequence robustness analysis.} We report the Abs Rel error ($\downarrow$) and Accuracy ($\delta < 1.25, \uparrow$) evaluated at increasing frame intervals (100 to 500). Our method (red) demonstrates minimal performance decay compared to FlashDepth and VideoDepthAnything, maintaining superior scale consistency even as the sequence length increases.}
  \label{fig:long_sequence}
  \vspace{-15pt}
\end{figure}

To further demonstrate the robustness of our method against scale drift, we conducted experiments on 100 selected scenes from the ScanNet dataset~\cite{dai2017scannet}, each comprising a continuous sequence of 500 frames. 
We compared our approach against two state-of-the-art baselines: FlashDepth~\cite{chou2025flashdepth} and VideoDepthAnything~\cite{video_depth_anything}. 
The evaluation was performed at incremental intervals of 100 frames. 
Crucially, to rigorously test global consistency, we re-calculated the sequence-wise scale and shift alignment at each evaluation step. 
As illustrated in Fig.~\ref{fig:long_sequence}, while extended sequence lengths generally introduce accumulated error, our method demonstrates significantly superior stability. 
The \textit{Ours} curve exhibits a much slower rate of degradation in both Absolute Relative Error (AbsRel) and Accuracy ($\delta < 1.25$) compared to the baselines. 
Notably, even at the 500-frame mark, our method preserves high accuracy and outperforms both FlashDepth and VDA by a clear margin. 
This empirically validates our method's effectiveness in maintaining scale consistency over long durations.

\section{Conclusion}
We identified that temporal inconsistency in monocular geometry models stems from latent feature fluctuations causing scale-shift drift. 
We introduced Dynamic Feature Normalization, a lightweight recurrent module that stabilizes these feature statistics over time. 
By finetuning only DyFN while freezing the pretrained backbone, our method preserves single-frame accuracy while achieving state-of-the-art temporal stability. 
Our results demonstrate superior performance, even on {long-duration sequences} where we mitigate the error accumulation that plagues other methods. 
This work offers a simple and efficient path for adapting static foundation models to continuous video streams.

\section{Acknowledgment}
This work was conducted during an internship at Voyager Research, DiDi Chuxing. The research was supported by the Hong Kong Research Grants Council (RGC) through the General Research Fund (Grants No. 17202422, 17212923, and 17215025), the Theme-based Research Scheme (Grant No. T45-701/22-R), and the Strategic Topics Grant (Grant No. STG3/E-605/25-N). Additionally, part of this research was conducted at the JC STEM Lab of Robotics for Soft Materials, funded by The Hong Kong Jockey Club Charities Trust. The authors would also like to thank Yikang Ding and Xin Kong for their valuable advice and insightful discussions throughout the course of this work.
\appendix
\renewcommand{\thesection}{S\arabic{section}} % S1, S2, S3...
\renewcommand{\thesubsection}{\thesection.\arabic{subsection}} % S1.1, S1.2...

\clearpage
\setcounter{page}{1}
\maketitlesupplementary
\section{Details of Loss Functions}
\label{sec: loss detail}

In addition to our proposed temporal alignment losses ($\mathcal{L}_{align}$ and $\mathcal{L}_{temp}$), we retain the original monocular supervision from MoGe~\cite{wang2025moge} to preserve single-frame geometric fidelity. 
The total monocular loss $\mathcal{L}_{MoGe}$ is composed of three terms:
\begin{equation}
    \mathcal{L}_{MoGe} = \mathcal{L}_{local} + \mathcal{L}_{normal} + \mathcal{L}_{mask},
\end{equation}
where $\mathcal{L}_{local}$, $\mathcal{L}_{normal}$, and $\mathcal{L}_{mask}$ supervise local geometry, surface normals, and validity masks, respectively.

\paragraph{Multi-scale local geometry loss ($\mathcal{L}_{local}$).} This term explicitly supervises local geometric structures. Given a ground-truth anchor point $\rm p_j$, we define a local spherical neighborhood $\mathcal{S}_j$ as:
\begin{equation}
    \mathcal{S}_j=\{i~|~||\rm p_i - \rm p_j||\leq r_j, i\in\mathcal{M}\}.
\end{equation}
Following MoGe, the radius $r_j$ is depth-adaptive, defined as $r_j=\alpha \cdot z_j \cdot \frac{\sqrt{W^2 + H^2}}{2\cdot f}$, where $z_j$ is the depth of $\rm p_j$, $f$ is the focal length, and $\alpha \in (0, 1)$ is a scalar controlling the neighborhood size relative to the image diagonal. 
Within each neighborhood, we solve for the optimal affine parameters $(s_j^*, t_j^*)$ to align predictions with the ground truth. We sample anchor sets $\mathcal{H}_\alpha$ across multiple scales $\alpha \in \{\frac{1}{4}, \frac{1}{16}, \frac{1}{32}\}$ and compute the accumulated error:
\begin{equation}
    \mathcal{L}_{local} = \sum_{\alpha}\sum_{j\in\mathcal{H}_\alpha}\sum_{i\in\mathcal{S}_j}\frac{1}{z_i}||s_j^*\rm \hat{p}_i + t^*_j - p_i||_1.
\end{equation}

\paragraph{Normal loss ($\mathcal{L}_{normal}$).} To enforce high-quality surface details, we minimize the angular error between predicted and ground-truth normals:
\begin{equation}
    \mathcal{L}_{normal} = \sum_{i\in \mathcal{M}}\angle (\hat{n}_i,n_i),
\end{equation}
where the predicted normal $\hat{n}_i$ is derived from the cross-product of adjacent vectors on the predicted point map grid, and $\angle (\cdot, \cdot)$ measures the angular difference.

\paragraph{Mask loss ($\mathcal{L}_{mask}$).} This loss is employed to identify valid geometric regions (e.g., suppressing sky or infinity in outdoor scenes). 
It is formulated as the mean squared error between the predicted mask $\hat{M}$ and the valid region label:
\begin{equation}
    \mathcal{L}_{mask}=||\hat{M} - ( 1 - M_{\text{inf}})||^2_2,
\end{equation}
where $M_{\text{inf}}$ denotes the infinity mask. 
During inference, $\hat{M}$ is binarized with a threshold of $0.5$.

\section{Details of Evaluation}
\label{sec: eval metric}
\paragraph{Evaluation Metrics.}
We adopt the Absolute Relative Error (AbsRel) and the inlier ratio $\delta_1$ as our primary metrics. Averaged over all valid pixels $\mathcal{M}$, these are defined as:
\begin{align}
    \text{AbsRel} &= \frac{1}{|\mathcal{M}|}\sum \frac{|d - \hat{d}|}{d}, \\
    \delta_1 &= \frac{1}{|\mathcal{M}|}\sum \mathbb{I}\left[\max\Big(\frac{d}{\hat{d}}, \frac{\hat{d}}{d}\Big) < 1.25\right],
\end{align}
where $d$ is the ground truth depth, $\hat{d}$ is the predicted depth (after alignment, if applicable), and $\mathbb{I}[\cdot]$ denotes the indicator function, which evaluates to 1 if the condition is met and 0 otherwise.

\paragraph{Evaluation Protocols.}
We employ three distinct protocols to evaluate different capabilities:

\noindent(1) \textbf{Metric Depth Protocol:} For models designed to predict absolute metric depth (e.g., DepthPro, MoGe-v2), we evaluate the raw predictions directly without any post alignment.

\noindent(2) \textbf{Video Depth Protocol (Global Alignment):} To evaluate temporal consistency, we align the entire predicted sequence using a \textbf{single} global transformation. Given predictions $\{\hat{\rm d}_j\}_{j=1}^L$ and ground truth $\{{\rm d_j}\}_{j=1}^L$, we solve for the optimal global scale $s^*$ and shift $t^*$ that minimize the error across all frames simultaneously:
\begin{equation}
    (s^*,t^*) = \mathop{\text{argmin}}_{s,t}\sum_{j=1}^L\sum_{i\in\mathcal{M}}\frac{1}{d^i_j}||s\hat{\rm d}_j^i + t - {\rm d}_j^i||_1.
\end{equation}
This global transformation is then applied uniformly to the sequence: $\{ {\rm d}^{align}_j\} = \{s^* \cdot {\hat{\rm d}}_j + t^*\}$. This protocol strictly penalizes scale drift over time.

\noindent(3) \textbf{Image Depth Protocol (Per-Frame Alignment):} To evaluate per-frame geometric quality in isolation, we align each frame independently. For each frame $j$, we compute specific parameters $(s^*_j, t^*_j)$:
\begin{equation}
    (s^*_j,t^*_j) = \mathop{\text{argmin}}_{s,t}\sum_{i\in\mathcal{M}}\frac{1}{d^i_j}||s\hat{\rm d}_j^i + t - {\rm d}_j^i||_1.
\end{equation}
The metrics are then computed on the individually aligned frames: $\{ {\rm d}^{align}_j\} = \{s^*_j \cdot {\hat{\rm d}}_j + t^*_j\}$.

\section{Dataset Configuration}
\label{sec: dataset config}
\subsection{Training Dataset}
\label{subsec: training dataset}
To finetune our newly designed DyFN module, we use seven different synthetic datasets which contain continuous frames and depth annotations.
Details are shown in the Tab~\ref{tab:datasets_details}.
Our training dataset contains total around 1M images and our main experiment are trained with the sequence length 12.
To increase the robustness of our model training, we randomly select stride from 1 to 5 to sample the continuous frames.
\begin{table}[ht]
    \centering
    \caption{Datasets used for training.}
    \label{tab:datasets_details}
    \begin{tabular}{lccc}
        \toprule
        Name & Domain & \# Frames & Weight \\
        \midrule
        IRS~\cite{wang2021irslargenaturalisticindoor} & Indoor & 101K & $20.1\%$\\
        PointOdyssey~\cite{zheng2023point} & Indoor & 79K & $27.8\%$ \\
        Dynamic Replica~\cite{karaev2023dynamicstereo} & Indoor & 143K & $17.4\%$ \\
        Spring~\cite{Mehl2023_Spring} & In-the-wild & 5K & $2.4\%$ \\
        MidAir~\cite{Fonder2019MidAir} & In-the-wild & 423K & $9.3\%$ \\
        KenBurns3D~\cite{Niklaus_TOG_2019} & In-the-wild & 76K & $5.6\%$ \\
        TartanAir~\cite{tartanair2020iros} & In-the-wild & 306K & $17.4\%$ \\
        
        \bottomrule
    \end{tabular}
\end{table}
\subsection{Evaluation Dataset}
\label{subsec: eval dataset}
\paragraph{Video \& Image Depth Estimation.}
We evaluate both video and image depth estimation performance using four diverse benchmarks. 
The details are as follows:
\begin{itemize}
    \item \textbf{Sintel~\cite{Butler:ECCV:2012}}. We utilize all 23 sequences for evaluation. 
    We evaluate directly at the original $1024\times436$ resolution without resizing.
    \item \textbf{ScanNet~\cite{dai2017scannet}}. We use the standard test split, comprising 100 scenes. We extract \textbf{90 continuous frames} per scene at a rate of 15 frames per second (FPS). 
    To handle the black borders resulting from calibration, we \textbf{follow} DepthCrafter~\cite{hu2024depthcrafter} and crop 8 pixels from the top and bottom edges, and 11 pixels from the left and right edges.
    \item \textbf{KITTI~\cite{kitti}}. We sample \textbf{110 frames} across all sequences in the official KITTI Depth split, maintaining the original frame rate.
    \item \textbf{Bonn~\cite{palazzolo2019iros}}. We selected 5 scenes from this dataset, each contributing 110 frames for evaluation.
\end{itemize}

\paragraph{Long Sequence Depth Estimation.}
To assess long-term stability and error accumulation, we adopt the same \textbf{ScanNetV2} test split. 
For this specific protocol, we extract \textbf{500 continuous frames} per scene, sampled at the depth camera's original frame rate. 
Furthermore, the same cropping strategy used for the short sequence evaluation is applied.

\section{Reconstruction Comparison}
\label{sec: recon}
\subsection{Reconstruction Algorithm}
\label{subsec: recon_alg}
% \paragraph{Pose Estimation for Scale-Shift Consistency.}
To rigorously evaluate the scale-shift consistency of our proposed model, we employ a geometric alignment protocol based on point correspondences. Given a sequence of $L$ continuous frames $\{\mathcal{I}_j\}_{j=1}^L$ and their predicted point clouds in the camera coordinate system $\{\mathcal{P}_j^{\text{cam}}\}_{j=1}^L$, the objective is to estimate the rigid transformation (pose) $\{R_j | t_j\}$ for the frame at timestamp $j$.

We select a set of reference frames $\mathcal{K} = \{j-1, j-5, j-21\}$ whose ground-truth poses are assumed known, providing their corresponding world coordinates $\{\mathcal{P}_k^{\text{world}}\}_{k \in \mathcal{K}}$. We first leverage PDCNet to establish reliable point correspondences, composing the matched 3D point pairs $\{\mathbf{p}_j^{\text{cam}}, \mathbf{p}_k^{\text{world}}\}_{k \in \mathcal{K}}$.

This set of correspondences is then used to solve for the optimal rigid transformation $\{R_j | t_j\}$ that aligns the predicted camera-centric point cloud $\mathcal{P}_j^{\text{cam}}$ into the global world coordinate system. To handle the inevitable noise and outliers present in the correspondences, we employ the {Random Sample Consensus (RANSAC)} algorithm. Within the RANSAC iterative loop, the rigid transformation $\{R_j, t_j\}$ is determined by solving the {Absolute Orientation Problem} (Procrustes problem). Specifically, we seek to minimize the squared error between the aligned source points and the target points:
\begin{equation}
    \min_{R_j, t_j} \sum_{m=1}^{N} \big\| R_j \mathbf{p}_{j, m}^{\text{cam}} + t_j - \mathbf{p}_{k, m}^{\text{world}} \big\|_2^2,
\end{equation}
where $N$ is the number of inlier correspondence pairs identified by RANSAC. The closed-form solution for the optimal rotation $R_j$ and translation $t_j$ is obtained efficiently using the Singular Value Decomposition (SVD) method applied to the centered cross-covariance matrix.
\subsection{Qualitative Comparison}
\label{subsec: recon_results}
\begin{figure*}[thbp]
  \vspace{-5pt}
  \centering
  \includegraphics[width=\textwidth]{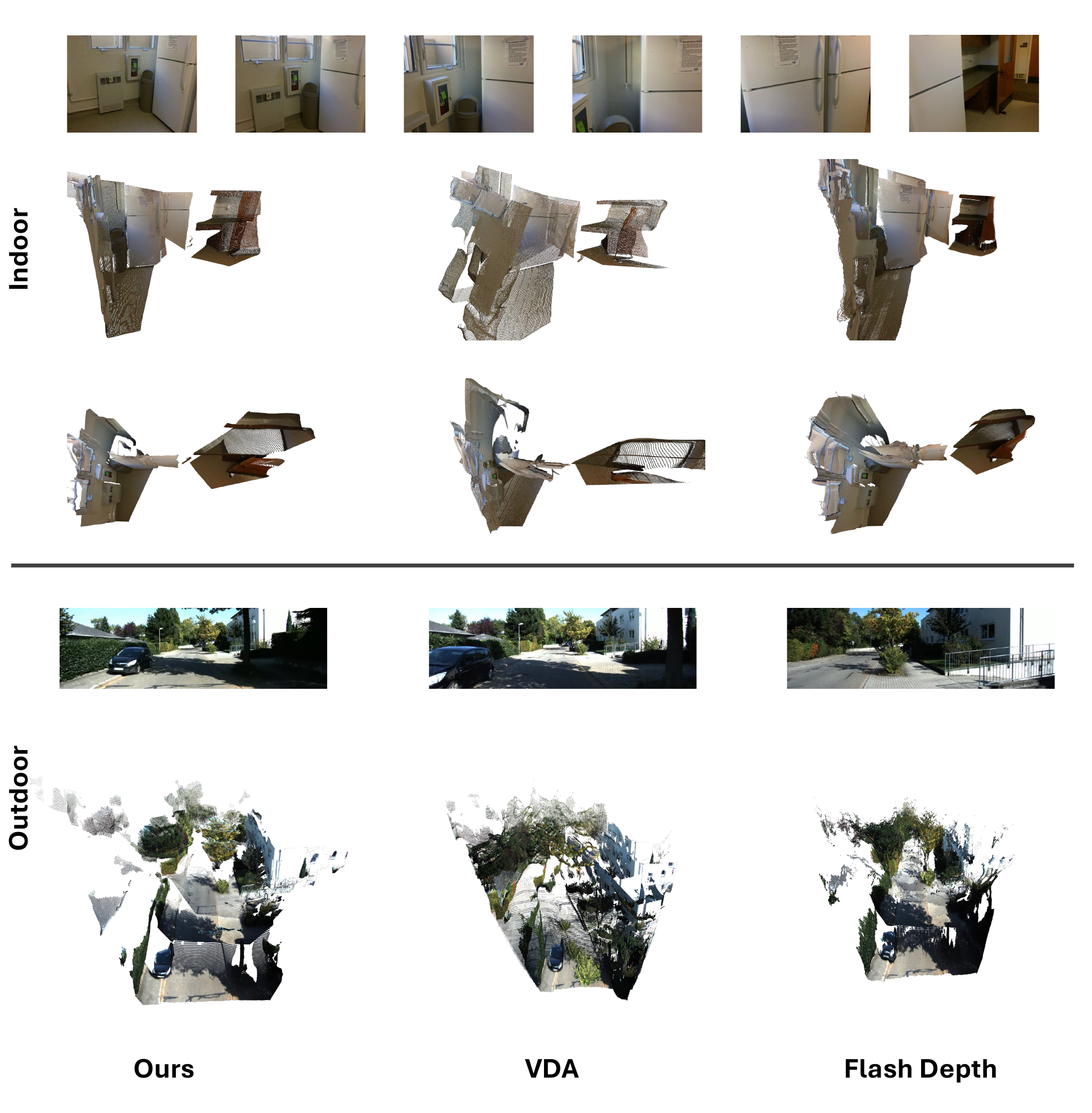}
  \caption{\textbf{Qualitative 3D Reconstruction Comparison in Diverse Scenes.} We present a qualitative comparison of 3D reconstruction results across challenging \textbf{indoor} and \textbf{outdoor} environments. Compared to key video depth baselines (VDA and FlashDepth), our method consistently demonstrates \textbf{superior geometric fidelity} and \textbf{enhanced temporal consistency}, resulting in noticeably more stable and accurate 3D structures.}
  \label{fig:supp_short}
  \vspace{-12pt}
\end{figure*}

\begin{figure*}[thbp]
  \vspace{-5pt}
  \centering
  \includegraphics[width=\textwidth]{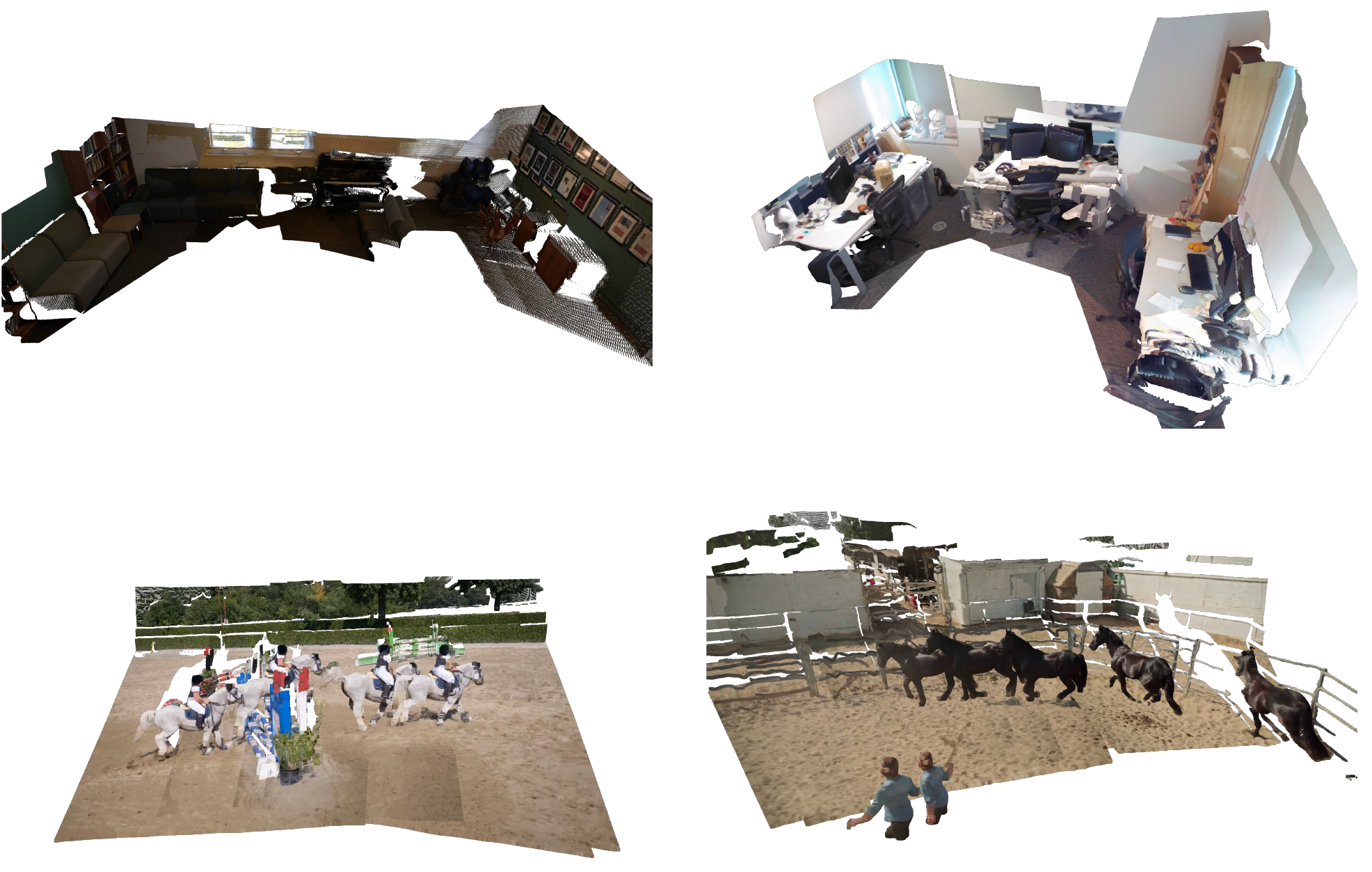}
  \caption{Qualitative results on long sequence reconstruction results (more than 500 frames) and dynamic reconstruction results.}
  \label{fig:supp_long_dynamic}
  \vspace{-12pt}
\end{figure*}

We utilize the robust reconstruction algorithm detailed in Section~\ref{subsec: recon_alg} for qualitative video reconstruction. Our results are presented in Figure~\ref{fig:supp_short} (short sequences) and Figure~\ref{fig:supp_long_dynamic} (long/dynamic sequences).

Figure~\ref{fig:supp_short} provides comparative results, demonstrating our approach's superior geometric consistency and clearer structural reconstruction in both indoor and outdoor scenes compared to baselines such as VideoDepthAnything (VDA) and FlashDepth. 
This highlights the immediate benefits of our dynamic feature stabilization.

Furthermore, Figure~\ref{fig:supp_long_dynamic} showcases our method's robust performance in complex scenarios. 
The results confirm sustained scale-shift consistency over \textbf{long-term sequences}, and crucially, illustrate the capability of our model to produce coherent geometric reconstructions even in the presence of dynamic scene elements.

\section{Training Length Influence}
\label{sec: training length}
To investigate the influence of training sequence length, we conduct an ablation study across four evaluation benchmarks, as detailed in Table~\ref{tab: len_ablation}. 
The results clearly demonstrate that varying the input frame length from 8 to 24 frames has a \textbf{negligible impact} on the final accuracy metrics across all datasets. 
This observation strongly confirms the inherent stability of our DyFN module and demonstrates that the necessary temporal alignment and scale-shift information are learned highly \textbf{efficiently}, even from relatively short sequence clips. 
This stability allows us to select 12 frames as the standard training length, optimizing the balance between computational efficiency and stable performance.

\begin{table}[t]
    \centering 
    % \resizebox{\linewidth}{!}{
    \setlength{\tabcolsep}{2pt}
    \adjustbox{width={\linewidth},keepaspectratio}{
    \begin{tabular}{ccccccccc}
        \bottomrule
        \multirow{2}{*}{} & \multicolumn{2}{c}{\textbf{Sintel }} & \multicolumn{2}{c}{\textbf{Scannet}} & \multicolumn{2}{c}{\textbf{KITTI}} & \multicolumn{2}{c}{\textbf{Bonn}} \\
    \cmidrule(lr){2-3} \cmidrule(lr){4-5} \cmidrule(lr){6-7} \cmidrule(lr){8-9}
     \textbf{Frame Length} & Abs Rel$\downarrow$ & $\delta < 1.25 \uparrow$ & Abs Rel$\downarrow$ & $\delta < 1.25 \uparrow$ & Abs Rel$\downarrow$ & $\delta < 1.25 \uparrow$ & Abs Rel$\downarrow$ & $\delta < 1.25 \uparrow$ \\
        \midrule
        8     & 0.182 & 73.3 & 0.072 & 96.2 & 0.064 & 97.3 & 0.043 & 98.4 \\
        12    & 0.180 & 73.0 & 0.073 & 96.6 & 0.062 & 97.3 & 0.044 & 98.4 \\
        16    & 0.181 & 73.4 & 0.072 & 96.2 & 0.064 & 97.0 & 0.044 & 98.4 \\
        24    & 0.182 & 73.3 & 0.071 & 96.3 & 0.067 & 97.6 & 0.045 & 98.4 \\
        \toprule
    \end{tabular}}
    \caption{\textbf{Ablation Study on Training Sequence Length.} The negligible variation in performance across different sequence lengths (8 to 24 frames) confirms the stability and efficient learning of temporal consistency in our method.}
    \label{tab: len_ablation}
    \vspace{-12pt}
\end{table}

\section{Implementation Details}
\label{sec: recurrent}
\begin{figure*}[thbp]
  \vspace{-5pt}
  \centering
  \includegraphics[width=\textwidth]{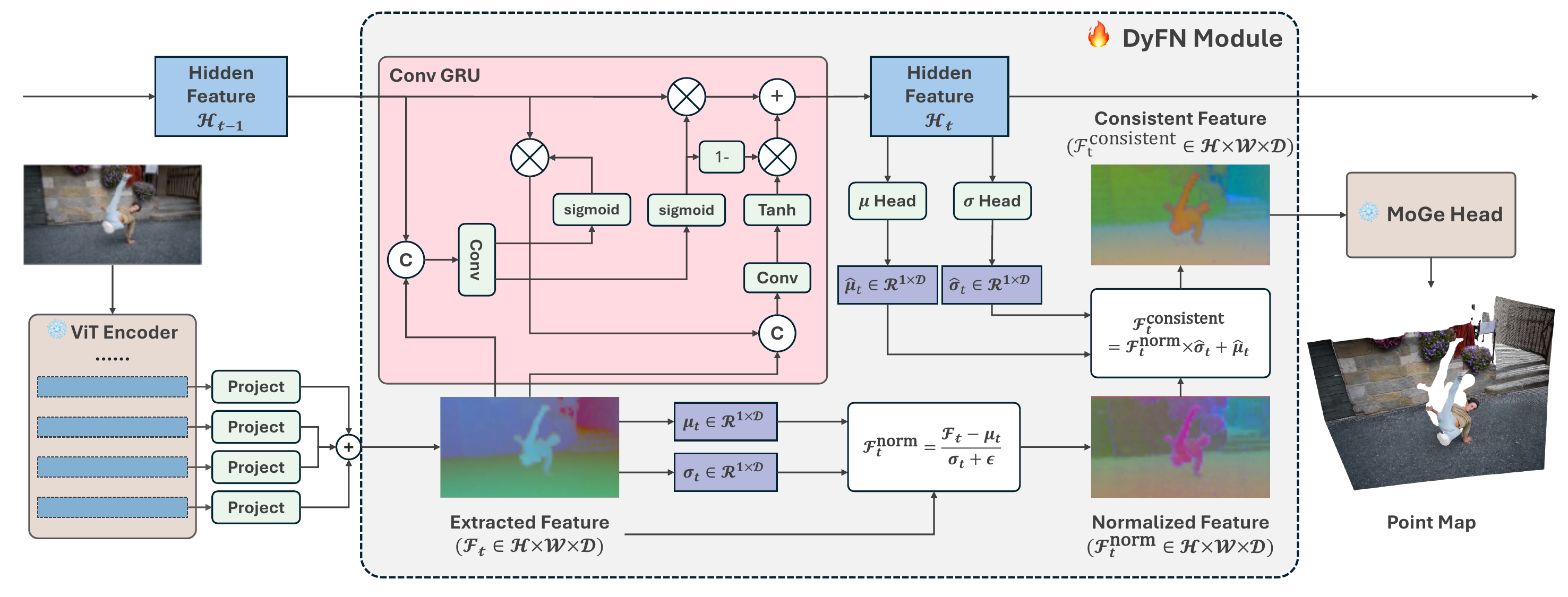}
  \caption{\textbf{Detailed network structures with ConvGRU.} We show the detailed structures when DyFN module use the ConvGRU as recurrent module to merge the historical information.}
  \label{fig:detailed_conGRU}
  \vspace{-12pt}
\end{figure*}

As illustrated in Figure~\ref{fig:detailed_conGRU}, our proposed method utilizes a \textbf{single ConvGRU} recurrent structure to model temporal dependencies and generate the hidden state. 
This targeted design leads to {superior parameter efficiency}: we only need to optimize the weights of the ConvGRU module, which constitutes a dramatically reduced parameter budget of \textbf{$2\%$ (approximately $5\text{M}$)} of the total network parameters. 
This is orders of magnitude lower than previous video-trained methods like DepthCrafter ($1422.8\text{M}$) and VideoDepthAnything ($381.8\text{M}$), allowing for highly efficient fine-tuning and deployment.

\section{Generalization Capability}
\label{sec: generalization}
To rigorously demonstrate the \textbf{generalization capability} of our proposed DyFN module, we integrate it into the \textbf{DepthAnythingV2 (DAv2)} framework. 
Unlike our primary Monocular Geometry Estimation (MGE) backbone, DAv2 is a standard monocular depth model designed to output \textbf{disparity}, rendering the MoGe-specific geometry losses ($\mathcal{L}_{\text{MoGe}}$) unsuitable. 
To adapt, we utilize two key supervision signals: the standard \textbf{scale-shift invariant loss} ($\mathcal{L}_{\text{ssi}}$) proposed in MiDaS~\cite{birkl2023midas} and our inter-frame temporal loss ($\mathcal{L}_{\text{temp}}$) defined in Equation~\ref{eq:temp_loss}. 
We adopt the same parameter-efficient fine-tuning strategy (freezing the backbone). 
As shown in Table~\ref{tab: dav2}, integrating the DyFN module dramatically boosts DAv2's performance across diverse domains. 
Specifically, the $\delta_1$ accuracy on Sintel improves substantially from $55.4$ to $\mathbf{63.0}$, and on KITTI, it rises sharply from $80.4$ to $\mathbf{92.9}$. 
This significant uplift validates that DyFN's mechanism for stabilizing feature statistics is broadly effective across different architectural types and output representations.

% \begin{table}[t]
%     \centering 
%     % \resizebox{\linewidth}{!}{
%     \setlength{\tabcolsep}{2pt}
%     \adjustbox{width={\linewidth},keepaspectratio}{
%     \begin{tabular}{ccccc}
%         \bottomrule
%         \multirow{2}{*}{} & \multicolumn{2}{c}{\textbf{Sintel }} & \multicolumn{2}{c}{\textbf{KITTI}} \\
%     \cmidrule(lr){2-3} \cmidrule(lr){4-5}
%      \textbf{Method} & Abs Rel$\downarrow$ & $\delta < 1.25 \uparrow$ & Abs Rel$\downarrow$ & $\delta < 1.25 \uparrow$ \\
%         \midrule
%         DAv2     & 0.367 & 55.4 & 0.140 & 80.4 \\
%         DAv2 + DyFN  & \textbf{0.303} & \textbf{63.0} & \textbf{0.090} & \textbf{92.9} \\
%         \toprule
%     \end{tabular}}
%     \caption{Quantitative results with the depth anything v2 backbone.}
%     \label{tab: dav2}
%     \vspace{-12pt}
% \end{table}
\begin{table}[htbp]
    \centering
    % \vspace{-10pt}
    \setlength{\tabcolsep}{1pt}
    % \resizebox{\linewidth}{!}{
    \adjustbox{width={\linewidth},keepaspectratio}{
    \begin{tabular}{lcccccccc}
        \bottomrule
        \multirow{2}{*}{} & \multicolumn{2}{c}{\textbf{Sintel }} & \multicolumn{2}{c}{\textbf{Scannet}} & \multicolumn{2}{c}{\textbf{KITTI}} & \multicolumn{2}{c}{\textbf{Bonn}} \\
    \cmidrule(lr){2-3} \cmidrule(lr){4-5} \cmidrule(lr){6-7} \cmidrule(lr){8-9}
     \textbf{Method} & Abs Rel$\downarrow$ & $\delta < 1.25 \uparrow$ & Abs Rel$\downarrow$ & $\delta < 1.25 \uparrow$ & Abs Rel$\downarrow$ & $\delta < 1.25 \uparrow$ & Abs Rel$\downarrow$ & $\delta < 1.25 \uparrow$ \\
        \midrule
        DAV2          & 0.367 & 55.4 & 0.135 & 82.2 & 0.140 & 80.4 & 0.106 & 92.1 \\
        FlashDepth    & 0.265 & 64.2 & 0.101 & 90.3 & 0.103 & 89.5 & \textbf{0.053} & \textbf{98.0} \\
        DAV2 + DyFN   & \textbf{0.242}  & \textbf{64.8} & \textbf{0.087}  & \textbf{93.2}  & \textbf{0.093} & \textbf{93.3} & \textbf{0.053} & {97.9} \\
        \midrule
        MoGe         & 0.216 & 65.3 & 0.117 & 84.7 & 0.076 & 96.0 & 0.074 & 95.5 \\
        MoGe + DyFN   & \textbf{0.180} & \textbf{73.0} & \textbf{0.073} & \textbf{96.6} & \textbf{0.062} & \textbf{97.3} & \textbf{0.044} & \textbf{98.4} \\
        % \catD{} CUT3R & 0.428  & 55.4 & 0.064  & 93.7 & 0.092 & 91.3 & 0.063 & 96.2 \\
        \toprule
    \end{tabular}}
    \vspace{-12pt}
    \caption{\footnotesize Qualitative results on streaming video depth estimation.}
    \label{tab: dav2}
    % \vspace{-10pt}
\end{table}

\section{Limitation and Future Work}
\label{subsec: limitation}
While our Dynamic Feature Normalization (DyFN) module successfully mitigates temporal inconsistencies and maintains superior scale-shift consistency across long sequences, its performance ceiling is fundamentally constrained. 
A primary limitation is that the achievable accuracy remains bounded by the \textbf{per-frame aligned geometric fidelity} of the underlying monocular depth backbone. 
Since DyFN operates by stabilizing the existing feature representation, it does not leverage the redundant information across multiple continuous frames to resolve fundamental monocular ambiguities. 
Consequently, our method cannot inherently improve the geometric accuracy of any single frame beyond the backbone's original capability.

Future work will focus on extending the DyFN framework to better harness the structural cues present in continuous frames. 
By integrating multi-frame information within the recurrent structure, we aim to push past the conventional limits of single-image depth estimation, significantly enhancing the geometric fidelity and ambiguity resolution capacity of the resulting depth predictions.
% \section{Metric Depth Estimation}
% \label{sec: metric}

\clearpage
{
    \small
    \bibliographystyle{ieeenat_fullname}
    \bibliography{main}

@String(CVPR= {IEEE Conf. Comput. Vis. Pattern Recog.})

@String(ICCV= {Int. Conf. Comput. Vis.})

@String(ECCV= {Eur. Conf. Comput. Vis.})

@String(ICLR = {Int. Conf. Learn. Represent.})

@String(CVPRW= {IEEE Conf. Comput. Vis. Pattern Recog. Worksh.})

@String(CVPR  = {CVPR})

@String(ICCV  = {ICCV})

@String(ECCV  = {ECCV})

@String(ICLR  = {ICLR})

@String(CVPRW= {CVPRW})

@String(IROS= {Proceedings of the {IEEE/RSJ} Conference on Intelligent Robots and Systems ({IROS})} )

@article{wang2025continuous,
  title={Continuous 3D Perception Model with Persistent State},
  author={Wang, Qianqian and Zhang, Yifei and Holynski, Aleksander and Efros, Alexei A and Kanazawa, Angjoo},
  journal={arXiv preprint arXiv:2501.12387},
  year={2025}
}

@inproceedings{ranftl2021vision,
  title={Vision transformers for dense prediction},
  author={Ranftl, Ren{\'e} and Bochkovskiy, Alexey and Koltun, Vladlen},
  booktitle={Proceedings of the IEEE/CVF international conference on computer vision},
  pages={12179--12188},
  year={2021}
}

@inproceedings{yin2021learning,
  title={Learning to Recover 3D Scene Shape from a Single Image},
  author={Yin, Wei and Zhang, Jianming and Wang, Oliver and Niklaus, Simon and Mai, Long and Chen, Simon and Shen, Chunhua},
  booktitle={Proceedings of the IEEE/CVF Conference on Computer Vision and Pattern Recognition},
  pages={204--213},
  year={2021}
}

@inproceedings{zheng2023point,
author = {Yang Zheng and Adam W. Harley and Bokui Shen and Gordon Wetzstein and Leonidas J. Guibas},
title = {PointOdyssey: A Large-Scale Synthetic Dataset for Long-Term Point Tracking},
booktitle = {ICCV},
year = {2023}
}

@article{bhat2023zoedepth,
  title={Zoedepth: Zero-shot transfer by combining relative and metric depth},
  author={Bhat, Shariq Farooq and Birkl, Reiner and Wofk, Diana and Wonka, Peter and M{\"u}ller, Matthias},
  journal={arXiv preprint arXiv:2302.12288},
  year={2023}
}

@inproceedings{yin2023metric3d,
  title={Metric3d: Towards zero-shot metric 3d prediction from a single image},
  author={Yin, Wei and Zhang, Chi and Chen, Hao and Cai, Zhipeng and Yu, Gang and Wang, Kaixuan and Chen, Xiaozhi and Shen, Chunhua},
  booktitle={Proceedings of the IEEE/CVF International Conference on Computer Vision},
  pages={9043--9053},
  year={2023}
}

@inproceedings{piccinelli2024unidepth,
  title={UniDepth: Universal monocular metric depth estimation},
  author={Piccinelli, Luigi and Yang, Yung-Hsu and Sakaridis, Christos and Segu, Mattia and Li, Siyuan and Van Gool, Luc and Yu, Fisher},
  booktitle={Proceedings of the IEEE/CVF Conference on Computer Vision and Pattern Recognition},
  pages={10106--10116},
  year={2024}
}

@article{ranftl2020towards,
  title={Towards robust monocular depth estimation: Mixing datasets for zero-shot cross-dataset transfer},
  author={Ranftl, Ren{\'e} and Lasinger, Katrin and Hafner, David and Schindler, Konrad and Koltun, Vladlen},
  journal={IEEE transactions on pattern analysis and machine intelligence},
  volume={44},
  number={3},
  pages={1623--1637},
  year={2020},
  publisher={IEEE}
}

@inproceedings{yang2024depth,
  title={Depth anything: Unleashing the power of large-scale unlabeled data},
  author={Yang, Lihe and Kang, Bingyi and Huang, Zilong and Xu, Xiaogang and Feng, Jiashi and Zhao, Hengshuang},
  booktitle={Proceedings of the IEEE/CVF Conference on Computer Vision and Pattern Recognition},
  pages={10371--10381},
  year={2024}
}

@article{yang2024depthv2,
  title={Depth anything v2},
  author={Yang, Lihe and Kang, Bingyi and Huang, Zilong and Zhao, Zhen and Xu, Xiaogang and Feng, Jiashi and Zhao, Hengshuang},
  journal={Advances in Neural Information Processing Systems},
  volume={37},
  pages={21875--21911},
  year={2024}
}

@inproceedings{ke2024repurposing,
  title={Repurposing diffusion-based image generators for monocular depth estimation},
  author={Ke, Bingxin and Obukhov, Anton and Huang, Shengyu and Metzger, Nando and Daudt, Rodrigo Caye and Schindler, Konrad},
  booktitle={Proceedings of the IEEE/CVF Conference on Computer Vision and Pattern Recognition},
  pages={9492--9502},
  year={2024}
}

@inproceedings{fu2024geowizard,
  title={Geowizard: Unleashing the diffusion priors for 3d geometry estimation from a single image},
  author={Fu, Xiao and Yin, Wei and Hu, Mu and Wang, Kaixuan and Ma, Yuexin and Tan, Ping and Shen, Shaojie and Lin, Dahua and Long, Xiaoxiao},
  booktitle={European Conference on Computer Vision},
  pages={241--258},
  year={2024},
  organization={Springer}
}

@article{chen2025video,
  title={Video Depth Anything: Consistent Depth Estimation for Super-Long Videos},
  author={Chen, Sili and Guo, Hengkai and Zhu, Shengnan and Zhang, Feihu and Huang, Zilong and Feng, Jiashi and Kang, Bingyi},
  journal={arXiv preprint arXiv:2501.12375},
  year={2025}
}

@article{ke2024video,
  title={Video Depth without Video Models},
  author={Ke, Bingxin and Narnhofer, Dominik and Huang, Shengyu and Ke, Lei and Peters, Torben and Fragkiadaki, Katerina and Obukhov, Anton and Schindler, Konrad},
  journal={arXiv preprint arXiv:2411.19189},
  year={2024}
}

@article{hu2024depthcrafter,
  title={Depthcrafter: Generating consistent long depth sequences for open-world videos},
  author={Hu, Wenbo and Gao, Xiangjun and Li, Xiaoyu and Zhao, Sijie and Cun, Xiaodong and Zhang, Yong and Quan, Long and Shan, Ying},
  journal={arXiv preprint arXiv:2409.02095},
  year={2024}
}

@article{chou2025flashdepth,
  title={FlashDepth: Real-time Streaming Video Depth Estimation at 2K Resolution},
  author={Chou, Gene and Xian, Wenqi and Yang, Guandao and Abdelfattah, Mohamed and Hariharan, Bharath and Snavely, Noah and Yu, Ning and Debevec, Paul},
  journal={arXiv preprint arXiv:2504.07093},
  year={2025}
}

@inproceedings{wang2024dust3r,
  title={Dust3r: Geometric 3d vision made easy},
  author={Wang, Shuzhe and Leroy, Vincent and Cabon, Yohann and Chidlovskii, Boris and Revaud, Jerome},
  booktitle={Proceedings of the IEEE/CVF Conference on Computer Vision and Pattern Recognition},
  pages={20697--20709},
  year={2024}
}

@article{zhang2024monst3r,
  title={Monst3r: A simple approach for estimating geometry in the presence of motion},
  author={Zhang, Junyi and Herrmann, Charles and Hur, Junhwa and Jampani, Varun and Darrell, Trevor and Cole, Forrester and Sun, Deqing and Yang, Ming-Hsuan},
  journal={arXiv preprint arXiv:2410.03825},
  year={2024}
}

@article{wang2025vggt,
  title={Vggt: Visual geometry grounded transformer},
  author={Wang, Jianyuan and Chen, Minghao and Karaev, Nikita and Vedaldi, Andrea and Rupprecht, Christian and Novotny, David},
  journal={arXiv preprint arXiv:2503.11651},
  year={2025}
}

@misc{oquab2023dinov2,
  title={DINOv2: Learning Robust Visual Features without Supervision},
  author={Oquab, Maxime and Darcet, Timothée and Moutakanni, Theo and Vo, Huy V. and Szafraniec, Marc and Khalidov, Vasil and Fernandez, Pierre and Haziza, Daniel and Massa, Francisco and El-Nouby, Alaaeldin and Howes, Russell and Huang, Po-Yao and Xu, Hu and Sharma, Vasu and Li, Shang-Wen and Galuba, Wojciech and Rabbat, Mike and Assran, Mido and Ballas, Nicolas and Synnaeve, Gabriel and Misra, Ishan and Jegou, Herve and Mairal, Julien and Labatut, Patrick and Joulin, Armand and Bojanowski, Piotr},
  journal={arXiv:2304.07193},
  year={2023}
}

@misc{shao2024learningtemporallyconsistentvideo,
      title={Learning Temporally Consistent Video Depth from Video Diffusion Priors}, 
      author={Jiahao Shao and Yuanbo Yang and Hongyu Zhou and Youmin Zhang and Yujun Shen and Vitor Guizilini and Yue Wang and Matteo Poggi and Yiyi Liao},
      year={2024},
      eprint={2406.01493},
      archivePrefix={arXiv},
      primaryClass={cs.CV},
      url={https://arxiv.org/abs/2406.01493}, 
}

@article{kitti,
  author = {Andreas Geiger and Philip Lenz and Christoph Stiller and Raquel Urtasun},
  title = {Vision meets Robotics: The KITTI Dataset},
  journal = {International Journal of Robotics Research (IJRR)},
  year = {2013}
}

@misc{wang2021irslargenaturalisticindoor,
      title={IRS: A Large Naturalistic Indoor Robotics Stereo Dataset to Train Deep Models for Disparity and Surface Normal Estimation}, 
      author={Qiang Wang and Shizhen Zheng and Qingsong Yan and Fei Deng and Kaiyong Zhao and Xiaowen Chu},
      year={2021},
      eprint={1912.09678},
      archivePrefix={arXiv},
      primaryClass={cs.CV},
      url={https://arxiv.org/abs/1912.09678}, 
}

@article{Niklaus_TOG_2019,
         author = {Simon Niklaus and Long Mai and Jimei Yang and Feng Liu},
         title = {3D Ken Burns Effect from a Single Image},
         journal = {ACM Transactions on Graphics},
         volume = {38},
         number = {6},
         pages = {184:1--184:15},
         year = {2019}
     }

@INPROCEEDINGS{Fonder2019MidAir,
author = {Michael Fonder and Marc Van Droogenbroeck},
title = {Mid-Air: A multi-modal dataset for extremely low altitude drone flights},
booktitle = {Conference on Computer Vision and Pattern Recognition Workshop (CVPRW)},
year = {2019},
month = {June}
}

@InProceedings{Mehl2023_Spring,
    author    = {Lukas Mehl and Jenny Schmalfuss and Azin Jahedi and Yaroslava Nalivayko and Andr\'es Bruhn},
    title     = {Spring: A High-Resolution High-Detail Dataset and Benchmark for Scene Flow, Optical Flow and Stereo},
    booktitle = {Proc. IEEE/CVF Conference on Computer Vision and Pattern Recognition (CVPR)},
    year      = {2023}
}

@article{tartanair2020iros,
  title =   {TartanAir: A Dataset to Push the Limits of Visual SLAM},
  author =  {Wang, Wenshan and Zhu, Delong and Wang, Xiangwei and Hu, Yaoyu and Qiu, Yuheng and Wang, Chen and Hu, Yafei and Kapoor, Ashish and Scherer, Sebastian},
  journal = {2020 IEEE/RSJ International Conference on Intelligent Robots and Systems (IROS)},
  year =    {2020}
}

@article{karaev2023dynamicstereo,
  title={DynamicStereo: Consistent Dynamic Depth from Stereo Videos},
  author={Nikita Karaev and Ignacio Rocco and Benjamin Graham and Natalia Neverova and Andrea Vedaldi and Christian Rupprecht},
  journal={CVPR},
  year={2023}
}

@inproceedings{bochkovskiydepth,
  title={Depth Pro: Sharp Monocular Metric Depth in Less Than a Second},
  author={Bochkovskiy, Alexey and Delaunoy, Ama{\"e}l and Germain, Hugo and Santos, Marcel and Zhou, Yichao and Richter, Stephan and Koltun, Vladlen},
  booktitle={The Thirteenth International Conference on Learning Representations},
year={2025}
}

@article{wang2025moge2,
  title={MoGe-2: Accurate Monocular Geometry with Metric Scale and Sharp Details},
  author={Wang, Ruicheng and Xu, Sicheng and Dong, Yue and Deng, Yu and Xiang, Jianfeng and Lv, Zelong and Sun, Guangzhong and Tong, Xin and Yang, Jiaolong},
  journal={arXiv preprint arXiv:2507.02546},
  year={2025}
}

@inproceedings{dai2017scannet,
  title={Scannet: Richly-annotated 3d reconstructions of indoor scenes},
  author={Dai, Angela and Chang, Angel X and Savva, Manolis and Halber, Maciej and Funkhouser, Thomas and Nie{\ss}ner, Matthias},
  booktitle={Proceedings of the IEEE conference on computer vision and pattern recognition},
  pages={5828--5839},
  year={2017}
}

@inproceedings{Butler:ECCV:2012,
title = {A naturalistic open source movie for optical flow evaluation},
author = {Butler, D. J. and Wulff, J. and Stanley, G. B. and Black, M. J.},
booktitle = {European Conf. on Computer Vision (ECCV)},
editor = {{A. Fitzgibbon et al. (Eds.)}},
series = {Part IV, LNCS 7577},
month = oct,
pages = {611--625},
year = {2012}
}

@inproceedings{guo2025depthcamerazeroshotmetric,
  title={Depth any camera: Zero-shot metric depth estimation from any camera},
  author={Guo, Yuliang and Garg, Sparsh and Miangoleh, S Mahdi H and Huang, Xinyu and Ren, Liu},
  booktitle={Proceedings of the Computer Vision and Pattern Recognition Conference},
  pages={26996--27006},
  year={2025}
}

@inproceedings{video_depth_anything,
  title={Video depth anything: Consistent depth estimation for super-long videos},
  author={Chen, Sili and Guo, Hengkai and Zhu, Shengnan and Zhang, Feihu and Huang, Zilong and Feng, Jiashi and Kang, Bingyi},
  booktitle={Proceedings of the Computer Vision and Pattern Recognition Conference},
  pages={22831--22840},
  year={2025}
}

@inproceedings{MegaDepthLi18,
  title={Megadepth: Learning single-view depth prediction from internet photos},
  author={Li, Zhengqi and Snavely, Noah},
  booktitle={Proceedings of the IEEE conference on computer vision and pattern recognition},
  pages={2041--2050},
  year={2018}
}

@inproceedings{wang2025moge,
  title={Moge: Unlocking accurate monocular geometry estimation for open-domain images with optimal training supervision},
  author={Wang, Ruicheng and Xu, Sicheng and Dai, Cassie and Xiang, Jianfeng and Deng, Yu and Tong, Xin and Yang, Jiaolong},
  booktitle={Proceedings of the Computer Vision and Pattern Recognition Conference},
  pages={5261--5271},
  year={2025}
}

@article{Hu_2024,
   title={Metric3D v2: A Versatile Monocular Geometric Foundation Model for Zero-Shot Metric Depth and Surface Normal Estimation},
   volume={46},
   ISSN={1939-3539},
   url={http://dx.doi.org/10.1109/TPAMI.2024.3444912},
   DOI={10.1109/tpami.2024.3444912},
   number={12},
   journal={IEEE Transactions on Pattern Analysis and Machine Intelligence},
   publisher={Institute of Electrical and Electronics Engineers (IEEE)},
   author={Hu, Mu and Yin, Wei and Zhang, Chi and Cai, Zhipeng and Long, Xiaoxiao and Chen, Hao and Wang, Kaixuan and Yu, Gang and Shen, Chunhua and Shen, Shaojie},
   year={2024},
   month=dec, pages={10579–10596} }

@misc{piccinelli2025unidepthv2universalmonocularmetric,
      title={UniDepthV2: Universal Monocular Metric Depth Estimation Made Simpler}, 
      author={Luigi Piccinelli and Christos Sakaridis and Yung-Hsu Yang and Mattia Segu and Siyuan Li and Wim Abbeloos and Luc Van Gool},
      year={2025},
      eprint={2502.20110},
      archivePrefix={arXiv},
      primaryClass={cs.CV},
      url={https://arxiv.org/abs/2502.20110}, 
}

@article{birkl2023midas,
      title={MiDaS v3.1 -- A Model Zoo for Robust Monocular Relative Depth Estimation},
      author={Reiner Birkl and Diana Wofk and Matthias M{\"u}ller},
      journal={arXiv preprint arXiv:2307.14460},
      year={2023}
}

@inproceedings{rombach2022high,
  title={High-resolution image synthesis with latent diffusion models},
  author={Rombach, Robin and Blattmann, Andreas and Lorenz, Dominik and Esser, Patrick and Ommer, Bj{\"o}rn},
  booktitle={Proceedings of the IEEE/CVF conference on computer vision and pattern recognition},
  pages={10684--10695},
  year={2022}
}

@article{sun2025unigeo,
  title={UniGeo: Taming Video Diffusion for Unified Consistent Geometry Estimation},
  author={Sun, Yang-Tian and Yu, Xin and Huang, Zehuan and Huang, Yi-Hua and Guo, Yuan-Chen and Yang, Ziyi and Cao, Yan-Pei and Qi, Xiaojuan},
  journal={arXiv preprint arXiv:2505.24521},
  year={2025}
}

@inproceedings{eftekhar2021omnidata,
  title={Omnidata: A Scalable Pipeline for Making Multi-Task Mid-Level Vision Datasets From 3D Scans},
  author={Eftekhar, Ainaz and Sax, Alexander and Malik, Jitendra and Zamir, Amir},
  booktitle={Proceedings of the IEEE/CVF International Conference on Computer Vision},
  pages={10786--10796},
  year={2021}
}

@inproceedings{he2022masked,
  title={Masked autoencoders are scalable vision learners},
  author={He, Kaiming and Chen, Xinlei and Xie, Saining and Li, Yanghao and Doll{\'a}r, Piotr and Girshick, Ross},
  booktitle={Proceedings of the IEEE/CVF conference on computer vision and pattern recognition},
  pages={16000--16009},
  year={2022}
}

@article{weinzaepfel2022croco,
  title={Croco: Self-supervised pre-training for 3d vision tasks by cross-view completion},
  author={Weinzaepfel, Philippe and Leroy, Vincent and Lucas, Thomas and Br{\'e}gier, Romain and Cabon, Yohann and Arora, Vaibhav and Antsfeld, Leonid and Chidlovskii, Boris and Csurka, Gabriela and Revaud, J{\'e}r{\^o}me},
  journal={Advances in Neural Information Processing Systems},
  volume={35},
  pages={3502--3516},
  year={2022}
}

@inproceedings{weinzaepfel2023croco,
  title={Croco v2: Improved cross-view completion pre-training for stereo matching and optical flow},
  author={Weinzaepfel, Philippe and Lucas, Thomas and Leroy, Vincent and Cabon, Yohann and Arora, Vaibhav and Br{\'e}gier, Romain and Csurka, Gabriela and Antsfeld, Leonid and Chidlovskii, Boris and Revaud, J{\'e}r{\^o}me},
  booktitle={Proceedings of the IEEE/CVF International Conference on Computer Vision},
  pages={17969--17980},
  year={2023}
}

@inproceedings{xie2017aggregated,
  title={Aggregated residual transformations for deep neural networks},
  author={Xie, Saining and Girshick, Ross and Doll{\'a}r, Piotr and Tu, Zhuowen and He, Kaiming},
  booktitle={Proceedings of the IEEE conference on computer vision and pattern recognition},
  pages={1492--1500},
  year={2017}
}

@article{podell2023sdxl,
  title={Sdxl: Improving latent diffusion models for high-resolution image synthesis},
  author={Podell, Dustin and English, Zion and Lacey, Kyle and Blattmann, Andreas and Dockhorn, Tim and M{\"u}ller, Jonas and Penna, Joe and Rombach, Robin},
  journal={arXiv preprint arXiv:2307.01952},
  year={2023}
}

@article{blattmann2023stable,
  title={Stable video diffusion: Scaling latent video diffusion models to large datasets},
  author={Blattmann, Andreas and Dockhorn, Tim and Kulal, Sumith and Mendelevitch, Daniel and Kilian, Maciej and Lorenz, Dominik and Levi, Yam and English, Zion and Voleti, Vikram and Letts, Adam and others},
  journal={arXiv preprint arXiv:2311.15127},
  year={2023}
}

@article{wan2025wan,
  title={Wan: Open and advanced large-scale video generative models},
  author={Wan, Team and Wang, Ang and Ai, Baole and Wen, Bin and Mao, Chaojie and Xie, Chen-Wei and Chen, Di and Yu, Feiwu and Zhao, Haiming and Yang, Jianxiao and others},
  journal={arXiv preprint arXiv:2503.20314},
  year={2025}
}

@inproceedings{tan2023temporal,
  title={Temporal attention unit: Towards efficient spatiotemporal predictive learning},
  author={Tan, Cheng and Gao, Zhangyang and Wu, Lirong and Xu, Yongjie and Xia, Jun and Li, Siyuan and Li, Stan Z},
  booktitle={Proceedings of the IEEE/CVF conference on computer vision and pattern recognition},
  pages={18770--18782},
  year={2023}
}

@inproceedings{gu2024mamba,
  title={Mamba: Linear-time sequence modeling with selective state spaces},
  author={Gu, Albert and Dao, Tri},
  booktitle={First conference on language modeling},
  year={2024}
}

@inproceedings{ballas2015delving,
  title={Delving deeper into convolutional networks for learning video representations},
  author={Ballas, Nicolas and Li, Yao and Pal, Chris and Courville, Aaron},
  booktitle={International Conference on Learning Representations (ICLR)},
  year={2016}
}

@article{lstm,
author = {Hochreiter, Sepp and Schmidhuber, J\"{u}rgen},
title = {Long Short-Term Memory},
year = {1997},
issue_date = {November 15, 1997},
publisher = {MIT Press},
address = {Cambridge, MA, USA},
volume = {9},
number = {8},
issn = {0899-7667},
url = {https://doi.org/10.1162/neco.1997.9.8.1735},
doi = {10.1162/neco.1997.9.8.1735},
journal = {Neural Comput.},
month = nov,
pages = {1735–1780},
numpages = {46}
}

@article{wang20243d,
  title={3d reconstruction with spatial memory},
  author={Wang, Hengyi and Agapito, Lourdes},
  journal={arXiv preprint arXiv:2408.16061},
  year={2024}
}

@inproceedings{cabon2025must3r,
  title={Must3r: Multi-view network for stereo 3d reconstruction},
  author={Cabon, Yohann and Stoffl, Lucas and Antsfeld, Leonid and Csurka, Gabriela and Chidlovskii, Boris and Revaud, Jerome and Leroy, Vincent},
  booktitle={Proceedings of the Computer Vision and Pattern Recognition Conference},
  pages={1050--1060},
  year={2025}
}

@inproceedings{murai2025mast3r,
  title={MASt3R-SLAM: Real-time dense SLAM with 3D reconstruction priors},
  author={Murai, Riku and Dexheimer, Eric and Davison, Andrew J},
  booktitle={Proceedings of the Computer Vision and Pattern Recognition Conference},
  pages={16695--16705},
  year={2025}
}

@inproceedings{yang2025fast3r,
  title={Fast3r: Towards 3d reconstruction of 1000+ images in one forward pass},
  author={Yang, Jianing and Sax, Alexander and Liang, Kevin J and Henaff, Mikael and Tang, Hao and Cao, Ang and Chai, Joyce and Meier, Franziska and Feiszli, Matt},
  booktitle={Proceedings of the Computer Vision and Pattern Recognition Conference},
  pages={21924--21935},
  year={2025}
}

@article{chen2025ttt3r,
  title={Ttt3r: 3d reconstruction as test-time training},
  author={Chen, Xingyu and Chen, Yue and Xiu, Yuliang and Geiger, Andreas and Chen, Anpei},
  journal={arXiv preprint arXiv:2509.26645},
  year={2025}
}

@InProceedings{palazzolo2019iros,
author = {E. Palazzolo and J. Behley and P. Lottes and P. Gigu\`ere and C. Stachniss},
title = {{ReFusion: 3D Reconstruction in Dynamic Environments for RGB-D Cameras Exploiting Residuals}},
booktitle = IROS,
year = {2019},
url = {https://www.ipb.uni-bonn.de/pdfs/palazzolo2019iros.pdf},
codeurl = {https://github.com/PRBonn/refusion},
videourl = {https://youtu.be/1P9ZfIS5-p4},
}
}


\end{document}